\PassOptionsToPackage{dvipsnames}{xcolor}

\documentclass[11pt]{article}

\usepackage[T1]{fontenc}
\usepackage[utf8]{inputenc}
\usepackage{silence}
\usepackage{microtype}
\usepackage{times}

\usepackage[]{acl}
\usepackage{graphicx}
\usepackage[most]{tcolorbox}
\usepackage{booktabs}
\usepackage{tabularx}
\usepackage{siunitx}
\usepackage{makecell}
\usepackage{array}
\usepackage{xspace}
\usepackage{listings} 
\usepackage{float}
\usepackage{caption}
\usepackage{graphbox}
\usepackage{multirow} 
\usepackage{ragged2e}

\newcolumntype{Y}{>{\arraybackslash}X}

\definecolor{diffadd}{RGB}{0, 100, 0}
\definecolor{diffremove}{RGB}{180, 0, 0}
\definecolor{codegray}{rgb}{0.5,0.5,0.5}

\lstdefinestyle{diffstyle}{
    basicstyle=\ttfamily\small,
    morecomment=[f][\color{diffadd}]{+},    
    morecomment=[f][\color{diffremove}]{-}, 
    commentstyle=\color{codegray},
    numbers=none,
    frame=none,
    keepspaces=true,
    breaklines=true,
}

\lstdefinelanguage{Julia}%
  {morekeywords={abstract,break,case,catch,const,continue,do,else,elseif,%
      end,export,false,for,function,immutable,import,importall,if,in,%
      macro,module,otherwise,quote,return,switch,true,try,type,typealias,%
      using,while},%
   sensitive=true,%
   alsoother={$},%
   morecomment=[l]\#,%
   morecomment=[n]{\#=}{=\#},%
   morestring=[s]{"}{"},%
   morestring=[m]{'}{'},%
}[keywords,comments,strings]%

\usepackage{newunicodechar}
\newunicodechar{λ}{\ensuremath{\lambda}}
\newunicodechar{σ}{\ensuremath{\sigma}}
\newunicodechar{α}{\ensuremath{\alpha}}
\newunicodechar{μ}{\ensuremath{\mu}}
\newunicodechar{θ}{\ensuremath{\theta}}
\newunicodechar{τ}{\ensuremath{\tau}}
\newunicodechar{φ}{\ensuremath{\phi}}
\newunicodechar{ξ}{\ensuremath{\xi}}
\newunicodechar{⊙}{\ensuremath{\odot}}
\newunicodechar{∑}{\ensuremath{\sum}}
\newunicodechar{∈}{\ensuremath{\in}}
\newunicodechar{−}{\ensuremath{-}}
\newunicodechar{π}{\ensuremath{\pi}}
\newunicodechar{̂}{\ensuremath{^\wedge}}

\newcommand{\scicodeqa}{\textsc{SciCoQA}\xspace}

\defcitealias{deepseek-coder-v2}{DeepSeek-AI}

\definecolor{specGray}{RGB}{108, 117, 125}
\definecolor{specLine}{RGB}{222, 226, 230}
\definecolor{specBg}{RGB}{248, 249, 250}

\newtcolorbox{promptbox}[1][]{%
  enhanced,
  breakable,
  colback=white,
  colframe=specLine,
  boxrule=0.4pt,
  arc=2pt,
  outer arc=2pt,
  left=8pt, right=8pt, top=2pt, bottom=8pt,
  toptitle=5pt, bottomtitle=5pt,
  colbacktitle=specBg,
  coltitle=specGray,
  fonttitle=\sffamily\footnotesize,
  before title={\MakeUppercase{Prompt} ~\textbar~ },
  title={},
  fontupper=\sffamily\small,
  #1
}

\usepackage{fontawesome5}

\title{\textsc{SciCoQA}: Quality Assurance for Scientific Paper--Code Alignment}

\author{
Tim Baumgärtner \textnormal{\and} Iryna Gurevych \\
Ubiquitous Knowledge Processing Lab (UKP Lab), Department of Computer Science, \\
TU Darmstadt and National Research Center for Applied Cybersecurity ATHENE
\\
\\
\begin{small}
\begin{tabular}{l@{\hspace{0.5em}}l@{\hspace{0.3em}}l}
\includegraphics[align=c, height=1em]{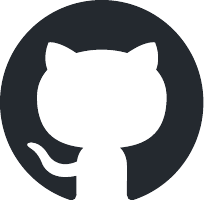} & \texttt{Code:} & \url{https://github.com/ukplab/scicoqa} \\
\includegraphics[align=c, height=1em]{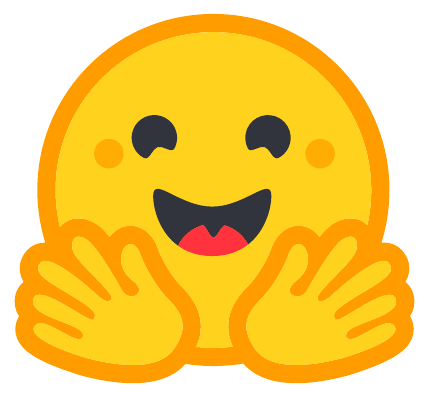} & \texttt{Data:} & \url{https://hf.co/datasets/ukplab/scicoqa} \\
\includegraphics[align=c, height=1em]{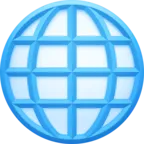} 
& \texttt{Blog:} & \url{https://ukplab.github.io/scicoqa}
\end{tabular}
\end{small}
}

\begin{document}
\maketitle

\begin{abstract}
Discrepancies between scientific papers and their code undermine reproducibility, a concern that grows as automated research agents scale scientific output beyond human review capacity. Whether LLMs can reliably detect such discrepancies has not been systematically measured. To this end, we present \scicodeqa, a dataset of $635$ paper-code discrepancies ($92$ real, $543$ synthetic) for this cross-modal verification task. Across $22$ evaluated models, even the best-performing LLMs, Gemini 3.1 Pro and GPT-5 Mini, detect only $46.7\%$ of real-world discrepancies, revealing a critical gap in automated scientific quality assurance. We construct \scicodeqa from GitHub issues and reproducibility papers, and propose a synthetic generation pipeline to scale beyond AI to Physics, Quantitative Biology, and other computational sciences. We further introduce a taxonomy of discrepancy types and categories to characterize the occurring mismatches. Our analysis shows that models particularly struggle with omitted paper details, long-context inputs, and papers outside their pre-training corpus.
\footnote{Data also available at: \url{https://tudatalib.ulb.tu-darmstadt.de/handle/tudatalib/4994}}
\end{abstract}

\section{Introduction}

The ``reproducibility crisis'' in AI and across science casts doubt on the reliability of research \citep{baker-reproducibility-crisis-2016, hutson-ai-reproducibility-crisis-2018}. 
To address this, the computational sciences have long recognized that a paper alone is insufficient and that publishing code, data, and instructions is a prerequisite to ensure experimental findings are reproducible \citep{wavelab-reproducible-buckheit-donoho-1995,reproducible-research-peng-etal-2011, raff-et-al-quantifying-ml-2019, improving-reproducibility-pineau-etal-2019}.
\begin{figure}[!ht]
    \centering
    \includegraphics[width=1\linewidth]{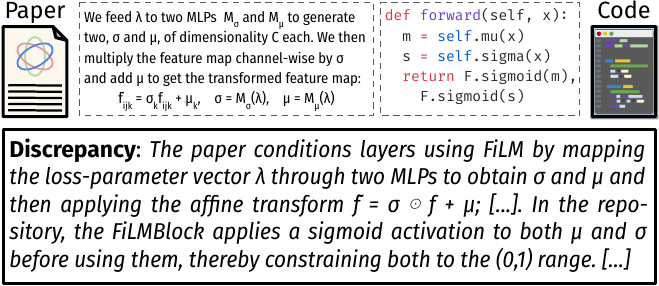}
    \caption{Example from \scicodeqa, showing a specific model implementation in the paper and its implementation in the code (simplified for readability). The paper's description and the code's implementation mismatch, creating a \textit{paper-code discrepancy}.}
    \label{fig:dataset-example}
\end{figure}
However, the availability of code guarantees neither reproducibility nor consistency with the scientific text (as showcased in Fig.~\ref{fig:dataset-example}). In practice, implementation details can diverge from their descriptions, introducing performance variations that go unreported \citep{deep-rl-that-matters-henderson-2018}. These discrepancies manifest in troubling ways: from ``mathiness,'' where equations simulate technical depth while actual gains stem from undocumented tricks \citep{troubling-trends-lipton-etal-2019}, to evaluation metrics that differ in implementation \citep{post-2018-call,marie-etal-2021-scientific}, rendering scientific comparisons invalid.

Reproducibility issues and paper-code inconsistencies are typically only detected during reproduction efforts and post-publication, resulting in a waste of resources and eroding trust in science. While checklists during peer review can encourage authors to provide more reproducible code \citep{improving-reproducibility-pineau-etal-2019,dodge-etal-2019-show}, ideally, reviewers check the correctness of the implementation and ensure reproducibility. However, with the rapid growth of submission numbers, reviewers are already under severe time pressure \citep{rogers-augenstein-2020-improve} and conducting intricate code reviews is time-consuming \citep{repeatability-in-research-collberg-proebsting-2016}.

Furthermore, the reliance on manual review is becoming increasingly impractical as science begins to scale via agentic systems that develop ideas, generate and execute code, and produce scientific articles autonomously \citep{boiko-etal-2023-auto-chemical,ai-scientist-lu-etal-2024,ai-researcher-tang-etal-2025,deepscientist-weng-etal-2025,lu-etal-2026-ai-scientist}, a trend recently exemplified by peer-reviewed acceptance to an ICLR workshop \citep{ai-scientist-v2-yamada-etal-2025}. While this direction has the potential to accelerate scientific discovery, it makes oversight increasingly challenging as humans cannot verify the rapidly expanding volume of output \citep{scalable-oversight-bowman-etal-2022}. Moreover, the faithfulness of these systems is not guaranteed as LLMs suffer from limited context \citep{liu-etal-2024-lost}, experience compounding errors \citep{faith-and-fate-dziri-etal-2023,gsm-symbolic-mirzadeh-etal-2025}, and struggle with independent self-correction \citep{tyen-etal-2024-llms,wu-etal-2024-large}. Consequently, ensuring the reliability of science at this scale demands automated tools capable of verifying that the code faithfully implements the methods reported in the scientific paper. 

To this end, we introduce \scicodeqa, the first benchmark for detecting discrepancies between scientific papers and their code as a cross-modal verification task. Benchmarking LLMs on this task allows us to measure their potential for automated quality assurance workflows in science. To construct our dataset, we leverage issues reported in code repositories and papers dedicated to reproducing existing work, such as those from reproducibility challenges and conference tracks. While these sources are realistic and of high quality, they are sparse and limited to CS and AI. Therefore, we scale the data to various computational science domains (e.g., Physics and Quantitative Biology) by generating modifications in codebases, creating discrepancies between the paper and its implementation. Finally, we benchmark open-weight and proprietary LLMs to determine whether current systems can be deployed for detecting paper-code discrepancies. Our evaluation reveals that the best-performing models, while precise in identifying discrepancies, suffer from insufficient recall. The best models in our evaluation detect only $46.7\%$ of real-world cases, highlighting the need to develop more effective models and systems for this critical task.

\section{Related Work}
\paragraph{Error Detection in Science} Recently, LLMs have been employed to detect errors in scientific papers. Early work created short CS papers with logical errors and prompted LLMs to identify the issues \citep{reviewer-gpt-liu-shah-2023}. However, their reliance on manually crafted papers limits scalability. Therefore, \citet{automatic-fail-dycke-etal-2025} and \citet{flaws-xi-etal-2025} propose dedicated pipelines that generate modifications to papers, invalidating their key claims. They evaluate whether the soundness score of generated reviews decreases or whether LLMs can spot these errors. Both find that LLMs struggle to detect most introduced errors. Besides constructing erroneous papers, recent research focuses on identifying real-world issues. Specifically, \citet{spot-son-etal-2025} and \citet{reviewing-critical-problems-zhang-abernethy-2025} use the authors' retraction notes from WithdrarXiv \citep{withdrarxiv-rao-etal-2024} and PubPeer to obtain faulty papers. While realistic, these notes are often unspecific (e.g., ``Theorem~2 is incorrect''). Recently, \citet{to-err-is-human-bianchi-etal-2025} developed a system to detect inconsistencies within the text of AI papers (e.g., incorrect calculations in tables or errors in equations) and then manually verified the detections. They found their system to be relatively precise but to struggle with recall, i.e., detecting all issues in a paper. They also noted an increasing trend of errors in scientific publications over the years, highlighting the importance of quality assurance in science.

\begin{figure*}[ht]
    \centering
    \includegraphics[width=1\linewidth,, trim={0 0.1cm 0 0.2cm},clip]{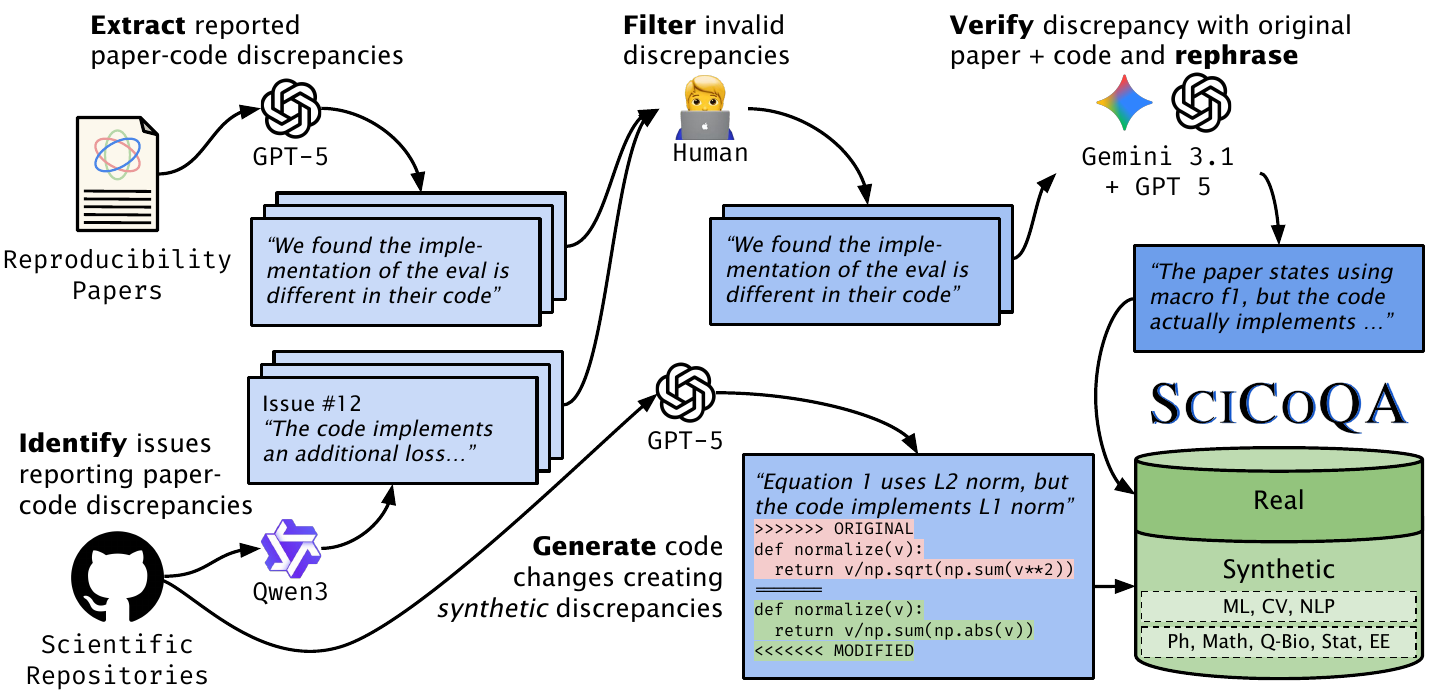}
    \caption{Overview of the data collection process of \scicodeqa. We source real-world data from reproducibility papers and GitHub issues. For the former, paper-code discrepancies are extracted from the paper with GPT-5, for the latter, issues are pre-filtered using Qwen3. Next, all candidates are manually filtered to remove any that do not fit our discrepancy definition. Finally, all paper-code discrepancies are verified with Gemini 3.1 and GPT-5. For synthetic data, we generate discrepancies using GPT-5 for AI and other computational domains.}
    \label{fig:overview}
\end{figure*}

While these works share a similar spirit to ours, i.e., identifying errors in science, they are limited by solely analyzing the paper. \scicodeqa extends beyond the scientific text and considers the code as well, enabling the detection of cross-modal error types, such as implementations that deviate from the paper description or omissions in the paper or the code that are crucial for understanding and reproducibility. Conceptually, our task can be viewed as a form of cross-modal claim verification \citep{thorne-etal-2018-fever,wadden-etal-2020-fact}, where the evidence source is a codebase rather than text. Furthermore, \scicodeqa utilizes constructed errors to scale the quantity and diversity in the computational science domain, while also providing challenging real-world discrepancies between paper and code.\looseness=-1

\paragraph{Code-Comment Inconsistency}
Our task is also related to Code-Comment Inconsistency (CCI) detection, where natural language comments that no longer match a piece of code need to be identified, e.g., after adding a feature to a function. Early CCI research relied on rule-based systems to identify outdated comments \citep{fragile-comments-ratol-robillard-2017}. Subsequently, bi-encoders and cross-encoders have been employed \citep{code-comment-siamese-rabbi-siddik-2020, code-comment-bert-steiner-zhang-2022}. \citet{deep-inconsistency-panthaplackel-2021} further extended CCI to dynamic settings, learning to detect inconsistency arising from code updates. Recently, LLMs have been deployed for both detection and rectification \citep{dau-etal-2024-docchecker, code-comment-llm-rong-etal-2025}. These methods deal with local and small inputs, typically taking a single function and an inline comment or docstring as input to detect the CCI. \scicodeqa is substantially more ambitious: it requires performing a global alignment by reasoning over a dense scientific paper and a long, multi-file code repository to find inconsistencies.

\paragraph{Science Automation} 
Automation in science is rapidly accelerating \citep{zheng-etal-2025-automation}. Early work targeted specific components of the research cycle, including ideation \citep{novel-ideas-llm-si-etal-2025,baek-etal-2025-researchagent}, literature review \citep{autosurvey-wang-etal-2024,surveyx-liang-etal-2025}, coding \citep{scicode-tian-etal-2024,researchcodeagent-gandhi-etal-2025}, and writing \citep{can-llm-feedback-liang-etal-2025,cycleresearcher-weng-etal-2025}. A recent trend involves generating entire code repositories from a paper \citep{paperbench-starace-etal-2025,paper2code-seo-etal-2025,scireplicatebench-xiang-etal-2025} or automating the scientific method end-to-end, giving rise to ``AI Scientists.'' However, for both these tasks, validating whether the generated code faithfully reflects the paper is challenging. For example, PaperBench \citep{paperbench-starace-etal-2025} relies on high-quality, but expensive, manual rubrics specific to each paper to verify implementation nuances. Others use generic LLM-judges without ground truth validation \citep{ai-researcher-tang-etal-2025}, evaluate whether the code runs \citep{deepscientist-weng-etal-2025,scireplicatebench-xiang-etal-2025}, or assign reproducibility scores based on execution outcomes \citep{corebench-siegel-etal-2024,hu-etal-2025-repro,replicationbench-ye-etal-2025}. Critically, these metrics can be misleading; a generated codebase may execute perfectly and achieve high performance while implementing a method that differs fundamentally from the paper's scientific description \citep{evaluating-ai-scientist-beel-etal-2025}. \scicodeqa addresses this gap by providing a ground truth dataset of paper-code discrepancies, enabling the development of robust quality assurance models that can verify the faithfulness of (generated) papers to (generated) code. More broadly, while code benchmarks in both software engineering \citep{swe-bench-jimenez-et-al-2023} and science \citep{scicode-tian-etal-2024} predominantly evaluate code generation, \scicodeqa evaluates the complementary task of code comprehension and verification against a scientific specification. Thus, \scicodeqa takes a step toward verification and reproducibility tooling essential for sustaining peer review integrity and science at scale \citep{wei-etal-2025-ai-imperative,woodruff-etal-2026-accelerating-science,you-etal-2026-preventing}. 

\section{\scicodeqa}\label{sec:scicodeqa}

The objective of the \scicodeqa task is to identify discrepancies between a paper and its code. We first define what constitutes a discrepancy:

\paragraph{Paper-Code Discrepancy} We define a paper-code discrepancy as a semantic conflict between the scientific method described in the publication and its actual implementation in the codebase, such that the code does not faithfully reproduce the reported method. This mismatch must be meaningful, implying a fundamental alteration to the scientific logic, experimental protocol, or mathematical formulation described in the text. These discrepancies manifest as three distinct types: \textit{Differences}, where the code implements a logic distinct from the paper's description (e.g., L1 vs. L2 normalization), \textit{Paper Omissions}, where the code includes critical components missing from the text, or \textit{Code Omissions}, where a step described in the paper is absent from the repository. We distinguish these discrepancies from engineering artifacts: We exclude bugs, as these are independent of the paper’s scientific description. Similarly, mismatches in default hyperparameters are not considered discrepancies if the code supports the paper's settings via configuration files or CLI arguments. Finally, we exclude trivial implementation details that are standard engineering practices typically omitted from scientific descriptions (e.g., adding noise to a denominator for numerical stability).

The paper-code discrepancies, according to our definition, can arise from distinct authorship (where the paper writer differs from the engineer implementing the code), simplifications made in the text for readability, or code updates or experiments that were not propagated to the manuscript.

\subsection{Data Collection}\label{sec:data-collection}
Fig.~\ref{fig:overview} provides an overview of our data collection. To obtain real-world instances of paper-code discrepancies, we draw from GitHub issues and reproducibility papers. Furthermore, synthetic discrepancies are generated in real scientific codebases. 

\paragraph{GitHub Issues}
We first identify repositories that reference a research paper in their homepage or description field on GitHub, restricted to projects published between 2020 and 2025 and repositories with at least one issue (see \S\ref{sec:github-crawl-details} for details).\footnote{The crawled repositories are not always the official implementation by the authors, but sometimes reproductions. They were published, for example, when there is no official implementation or the reimplementation uses a different framework. We still consider these repositories, as researchers might use them to base their experiments on.} For each, we crawl all associated issues, yielding $1{,}890$ repositories with a total of $10{,}636$ issues. To process this large volume efficiently, we automatically classify issues with Qwen3 4B Thinking \citep{yang-etal-qwen3-2025}, prompting the model to determine whether an issue reports a paper-code discrepancy or not, resulting in $232$ candidates (prompt in \S\ref{sec:prompts-github-discrepancy-extraction}). All candidates are subsequently manually filtered to ensure they meet our definition of a paper-code discrepancy, yielding $59$ discrepancies.

\begin{figure*}[!ht]
    \centering
    \includegraphics[width=1\linewidth, trim={0 0.35cm 0 0.25cm},clip]{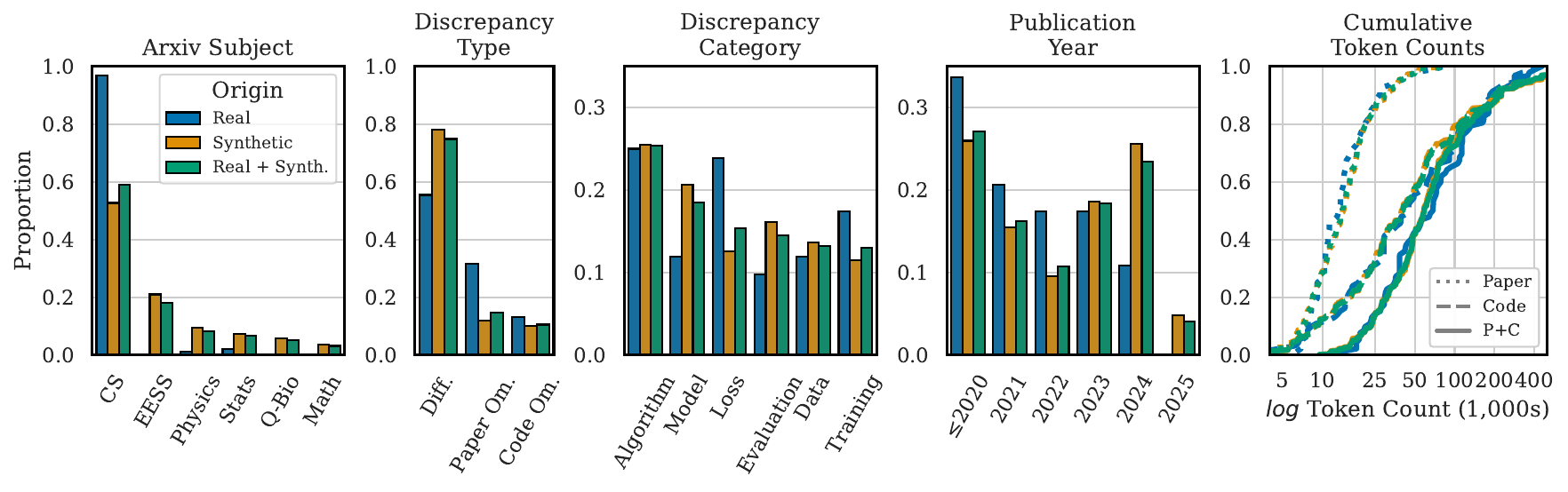}
    \caption{Analysis of the real (blue), synthetic (orange), and combined (green) discrepancy data. The y-axis shows the proportion of the data. The arXiv subjects stand for Computer Science (CS), Electrical Engineering and Systems Science (EESS), Physics, Statistics (Stats), Quantitative Biology (Q-Bio), and Mathematics (Math). The \textit{Discrepancy Category} chart shows only the distribution over computer science papers.}
    \label{fig:discrepancy-dataset-analysis}
\end{figure*}

\paragraph{Reproducibility Papers}
We collect reproducibility reports from the ML Reproducibility Challenge \citep{iclr-reproducibility-pineau-etal-2019,neurips-reproducibility-2019,ml-reproducibility-2021,ml-reproducibility-2022}, which invites participants to reproduce papers from leading ML, CV, and NLP conferences (e.g., NeurIPS, ICML, ICLR, CVPR, ICCV, ACL, EMNLP) and papers from the reproducibility tracks of SIGIR and ECIR from 2020–2025. We retain only reports that reproduce a single, open-access paper with an open-source implementation, resulting in $171$ reproducibility papers. From each report, we extract mentions of paper-code discrepancies using GPT-5, resulting in $132$ candidates (prompt in \S\ref{sec:prompts-github-reproducibility-report-extraction}). All extracted discrepancies are manually verified to confirm adherence to our definition, yielding $65$ discrepancies.

\paragraph{Validation and Phrasing} Finally, all manually filtered discrepancies are validated by Gemini 3.1 Pro and GPT-5, checking the discrepancy’s existence given the original paper, codebase, and the GitHub issue or the description extracted from the reproducibility paper. For the cases where the LLMs disagreed, we manually verified the discrepancy. During this step, the LLMs are further tasked with generating a standardized discrepancy description. We prompt the models to output 3-8 sentences and to state what the paper describes, what the code implements, and where the difference is. We use these generated descriptions as the ground truth to ensure the same format and level of verbosity for paper-code discrepancies (prompt in \S\ref{sec:prompts-github-discrepancy-verification} and \S\ref{sec:prompts-github-reproducibility-report-verification}). We use two state-of-the-art LLMs to perform the validation and rephrasing to prevent potential bias in the selection and phrasing (see \S\ref{sec:gemini-vs-gpt-ground-truth} for an analysis). The verification step ensures adherence to the discrepancy definition, acting as a high-precision filter. Therefore, we retain only discrepancies that can be verified given the explicit evidence, and prioritize the precision and objectivity of the ground truth. In total, we obtain $92$ discrepancies from $72$ papers, where $42$ discrepancies originate from GitHub issues and $50$ from reproducibility papers.

\paragraph{Synthetic Data} To scale the data in size and beyond the CS/AI domain, we employ synthetic data generation. Specifically, based on our GitHub crawl, we sample repositories linked to arXiv papers and those with permissive code licenses (i.e., MIT, Apache 2.0, CC-BY, BSD), as we will be redistributing their code partially. We randomly select $102$ repositories where the paper's arXiv subject is CS (balanced over Machine Learning (ML, cs.LG), Computer Vision (CV, cs.CV), and Natural Language Processing (NLP, cs.CL)), and $102$ for non-CS papers.\footnote{We manually checked the repositories and removed codebases from our sample that were not suitable, such as repositories for survey papers or with very few files.} Next, we prompt GPT-5 with the paper and code to generate five code diffs according to our discrepancy definition (prompt in \S\ref{sec:prompts-synthetic-discrepancy-generation}). We then sample up to three of these changes, ensuring that they do not manipulate the same file and that the generated diff can be found by exact match in the original code. Finally, we add $13$ additional discrepancies that are labeled as Paper Omissions as we later find these to be most critical (see \S\ref{sec:data-analysis} and \S\ref{sec:results}). This yields a total of $543$ discrepancies across the $204$ papers, of which $286$ discrepancies are from the CS domain, and $257$ are from other computational domains, such as Electrical Engineering and Systems Science ($114$), Physics ($51$), or Statistics ($40$).

Example data, including discrepancy type and category annotations, are shown in Table~\ref{tab:dataset-examples}. Examples from the synthetic subset, including the code changes, are shown in Table~\ref{tab:synthetic-examples}. We release \scicodeqa publicly under CC-BY-4.0.

\subsection{Data Analysis}\label{sec:data-analysis}
\paragraph{Domain} Fig.~\ref{fig:discrepancy-dataset-analysis} analyzes several dimensions of \scicodeqa. While our real-world data is mostly from computer science (specifically from AI and its subdomains such as ML, CV, and NLP), the synthetic data contains papers and code from Electrical Engineering and Systems Science, Physics, Statistics, Quantitative Biology, and Mathematics, making the data diverse across computational sciences.

\paragraph{Taxonomy} We further analyze the discrepancy type (as defined in \S\ref{sec:scicodeqa}) by annotating each discrepancy in the real set as \textit{Difference}, \textit{Paper Omission}, or \textit{Code Omission}. For the synthetic data, we provide type definitions in the prompt and generate the label along with the discrepancy. We find that $55\%$ of discrepancies in the real data are Differences, which is also the majority class in the synthetic data, accounting for $78\%$ of the data. Beyond the discrepancy type, we analyze the affected component within the research pipeline. We introduce a taxonomy of six discrepancy categories, namely: \textit{Algorithm}, \textit{Model}, \textit{Loss}, \textit{Evaluation}, \textit{Data}, and \textit{Training}. Table~\ref{tab:taxonomy} provides an overview of our proposed taxonomy. We apply this taxonomy to real-world data and the CS subset of synthetic data, excluding the other domains where these ML-specific concepts may not always apply cleanly. Similar to the discrepancy type annotation, we manually label the real instances and automatically generate labels for the synthetic ones. In the real data, discrepancies in the Algorithm and Loss categories dominate ($25\%$ and $24\%$), whereas in the synthetic data Algorithm is the largest category ($26\%$) and Model is the second-largest ($21\%$). 

\begin{table}[t]
\centering
\small
\begin{tabularx}{\columnwidth}{@{}llX@{}}
\toprule
& Label & Definition \\
\midrule
\multirow{6}{*}{\rotatebox[origin=c]{90}{\textit{Type}}}
& Difference & The code implements logic distinct from the paper's description \\
& Paper Omission & The code includes critical components not described in the paper \\
& Code Omission & A step described in the paper is absent from the code \\
\midrule
\multirow{10}{*}{\rotatebox[origin=c]{90}{\textit{Category}}}
& Algorithm & Changes to step order, operations, or core logic \\
& Model & Architectural or weight initialization deviations \\
& Loss & Alterations to loss definitions or terms \\
& Evaluation & Modifications to evaluation logic, metrics, or scripts \\
& Data & Dataset usage, preprocessing, augmentation, or filtering \\
& Training & Changes to the learning process, schedule, or optimization \\
\bottomrule
\end{tabularx}
\caption{The \scicodeqa discrepancy taxonomy. \textit{Types} describe the structural nature of the mismatch; \textit{Categories} describe which component of the research pipeline is affected.}
\label{tab:taxonomy}
\end{table}

\paragraph{Input Length} Finally, we visualize the token count distributions of the papers and codebases. The medians are $14{,}200$ tokens for papers, $39{,}272$ for codebases, and $56{,}903$ when concatenated. Further, $73$ out of $276$ papers have more than $100k$ tokens combined with their codebase, making \scicodeqa a challenging task for measuring the models' long-context abilities. In the appendix, we analyze the dataset by programming languages and the synthetic code diffs (\S\ref{sec:analysis-programming-language} and \S\ref{sec:analysis-synthetic-code}).

\section{Experiments}
Given a paper and its code, we prompt the model to generate a list of discrepancies between the two (prompt in \S\ref{sec:prompt-discrepancy-prediction}). We then parse the model output into individual discrepancies. The generation prompt contains the same instructions as those used to construct the ground-truth discrepancies. For implementation details, we refer to \S\ref{sec:implementation-details}. We further ablate the discrepancy prediction experiment by providing only the code as input. This allows us to quantify the contribution of the paper, distinguishing between discrepancies that require cross-modal reasoning and those that can be inferred from the code (e.g., through comments or readmes) or from the model's parametric knowledge of the paper.

\paragraph{Evaluation} We employ LLM-as-a-Judge \citep{llm-as-a-judge-zheng-et-al-2023} to evaluate whether the predicted discrepancy matches the reference. Inspired by \citet{wei-etal-2025-equibench}, we use a reasoning model, specifically GPT-OSS 20B \citep{gpt-oss-koch-et-al-2025} (prompt in \S\ref{sec:prompts-discrepancy-evaluation}), because it is open-weight, enabling reproducibility, and offers a favorable speed-performance trade-off. For the real data where two ground truth descriptions exist (generated by Gemini 3.1 Pro and GPT-5), we consider matches against either as correct. To verify the evaluation setup, we annotate the predictions from $20$ discrepancies, where $10$ originate from GitHub and $10$ from reproducibility papers. To streamline our evaluation, we compute the top three most similar predicted discrepancies per model using EmbeddingGemma \citep{schechter-etal-embeddinggemma-2025}. We then manually assess whether each prediction matches the reference discrepancy. In total, this yields $1{,}039$ annotations, of which $143$ predictions match the reference. On this data, the LLM judge achieves an F1 score of $87.5\% \pm 1.1$, showing strong alignment with our annotations.\footnote{We observe a slight variance in predictions despite a fixed seed in vLLM; we therefore report the mean and standard deviation over five runs.} 
We further validate the setup by running Qwen3 32B as an alternative judge. We find a Cohen's Kappa of $\kappa=0.838$ between them and an F1 score of $91.9\%$. However, since the F1 score of Qwen3 32B against our human evaluation is only $70.0\%$, we retain GPT-OSS 20B as a judge. While the evaluation is a binary classification (reference and predicted discrepancies match or do not match), we refer to the performance as \textit{recall}, as the predictions may contain unannotated, but valid, discrepancies, and we only evaluate whether the annotated ones are detected.

\paragraph{Models}
We evaluate several state-of-the-art model families, including commercial and open-weight models, as well as reasoning, instruction-tuned, and code-specific variants, specifically: GPT-5, Gemini, GPT-OSS, Qwen3, DeepSeek R1, Nemotron, and Mistral (exact model variants in Table~\ref{tab:model-version}).

\begin{figure*}[ht]
    \centering
    \includegraphics[width=0.975\linewidth, trim={0 0.28cm 0 0.25cm},clip]{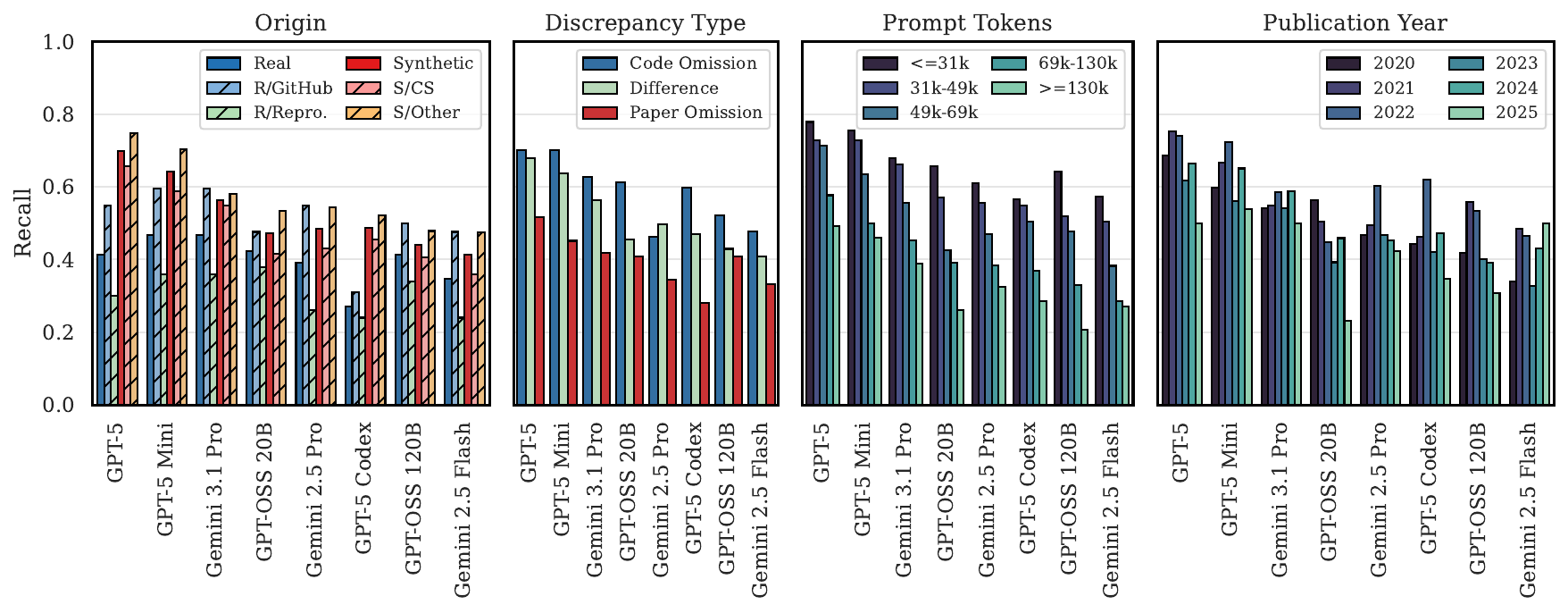}
    \caption{Results of the top 8 best-performing models (sorted by average recall on the real and synthetic data) on the discrepancy dataset by different analyses. From left to right: we analyze the origin of the discrepancy, the type of discrepancy, the number of tokens in the prompt, and the publication year of the paper.}
    \label{fig:discrepancy-results}
\end{figure*}

\section{Results}\label{sec:results}
Fig.~\ref{fig:discrepancy-results} reports the recall of the top-performing models (full results in Table~\ref{tab:detailed-results}). Overall, Gemini 3.1 Pro and GPT-5 Mini perform best, achieving a recall of $46.7\%$, leaving considerable room for improvement. On the synthetic data, GPT-5 detects $70.0\%$ of discrepancies. Crucially, we observe a strong correlation ($r=0.94$) between recall on real and synthetic data, validating the synthetic subset as a reliable proxy for ranking models (see \S\ref{sec:real-vs-synthetic-recall}). While general scaling laws can be observed (i.e., models trained with more compute and data perform better), GPT-5 Mini and GPT-OSS 20B are exceptions in our task, slightly outperforming their larger variants. We attribute this partly to the self-preference bias of the judge for GPT-OSS, rating its own outputs better \citep{llm-evaluators-recognize-panickssery-etal-2024}, and partly to the model's verbosity. GPT-5 Mini and GPT-OSS 20B on average make $5.3$ and $5.8$ predictions per paper, while GPT-5 and GPT-OSS 120B make only $4.6$ and $5.0$ predictions, giving the smaller variants a higher chance to match the discrepancy at the expense of precision. Additionally, we find GPT-5 Codex to be inferior, despite being a larger model than GPT-5 Mini (Codex is based on GPT-5). While Codex is generally superior in code generation, for \scicodeqa, code and natural language understanding are both crucial, and we conjecture that the general instruction-following and reasoning abilities of GPT-5 and GPT-5 Mini are more helpful than specialized coding knowledge.\looseness=-1

\paragraph{Origin and Type}
Analyzing detection rates reveals a clear hierarchy: Code Omissions are the easiest to detect, followed by Differences, while Paper Omissions are the most challenging. This order also explains the performance gap between data sources. Discrepancies from GitHub are easier to detect because they primarily consist of Differences ($71.4\%$) and Code Omissions ($19.0\%$). In contrast, discrepancies from reproducibility papers are more challenging as they are often dominated by Paper Omissions ($50\%$). The challenging nature of Paper Omissions likely stems from the asymmetry between modalities: the paper acts as a specification, where all described components must exist in the code. Code Omissions and Differences benefit from this explicit grounding. Conversely, the code contains many implementation details not required in the paper; thus, detecting Paper Omissions requires the harder task of distinguishing deviations from permissible engineering artifacts without a reference in the paper.

\paragraph{Input Length} We further analyze the performance by the number of input tokens.\footnote{ \S\ref{sec:truncation-analysis} conducts a further performance analysis by the position of the relevant code file in the prompt.} We split the dataset by prompt length into five approximately equally sized bins. We find a consistent pattern where the longer the input, the lower the performance becomes, consistent with findings in prior work \citep{zhang-etal-2024-bench,levy-etal-2024-task}.
This sensitivity to input length also suggests an explanation for the performance gap between domains in the synthetic data: models struggle more with CS papers, which is likely driven by the larger repository size of CS vs. non-CS repositories (median: $53k$ vs. $31k$ tokens). This increased volume creates a larger search space, making the task more challenging. Unlike long-context evaluations that primarily test retrieval of isolated facts from extended contexts \citep{needle-kamradt-2023,ruler-hsieh-etal-2024}, \scicodeqa requires cross-document reasoning over real scientific papers and multi-file codebases to identify semantic mismatches, making it a natural testbed for long-context abilities (similar to the question answering task of \citet{repoqa-liu-etal-2024}).\footnote{The synthetic data generation can further be used to inject discrepancies at controlled positions in the codebase, enabling targeted analysis of context-length effects.}

\paragraph{Publication Year} Lastly, we split the data by publication year. Among the top models, the one with the most recent knowledge cutoff is the Gemini family, which was trained using data up to January 2025 (other models have cutoffs in 2024). Therefore, the 2025 split of our data is largely outside the pre-training data, although Gemini models have partial coverage. Most models (except Gemini 2.5 Flash) perform worst on the most recent data. This suggests that models benefit from the inclusion of specific papers and codebases in their pre-training sets, lending support to concerns regarding data contamination discussed in the literature \citep{gpt4-openai-2023, benchmark-cheater-zhuo-etal-2023, sainz-etal-2023-nlp}. This further emphasizes the importance of our synthetic data generation pipeline, which enables us to continuously update our dataset with uncontaminated data that is not part of the pre-training data of future models.

\paragraph{Open-Weight Models} Beyond the top-performing models, we find the other open-weight models (Qwen3, Nemotron, DeepSeek R1, Mistral) severely limited. Nemotron 49B and Qwen3 30B Coder perform best, but only achieve an average recall on the combined data of $23.9\%$ and $23.5\%$, respectively.

In summary, \scicodeqa provides a significant challenge for state-of-the-art LLMs. We find the recall of all models to be limited; the best-performing models can only detect $46.7\%$ of all discrepancies in the real-world data. Consequently, these systems cannot yet be relied upon to ensure the faithfulness of scientific publications to their codebases.

\subsection{Code-Only Ablation}
\begin{figure}[tb]
    \centering
    \includegraphics[width=\linewidth, trim={0 0.28cm 0 0.26cm},clip]{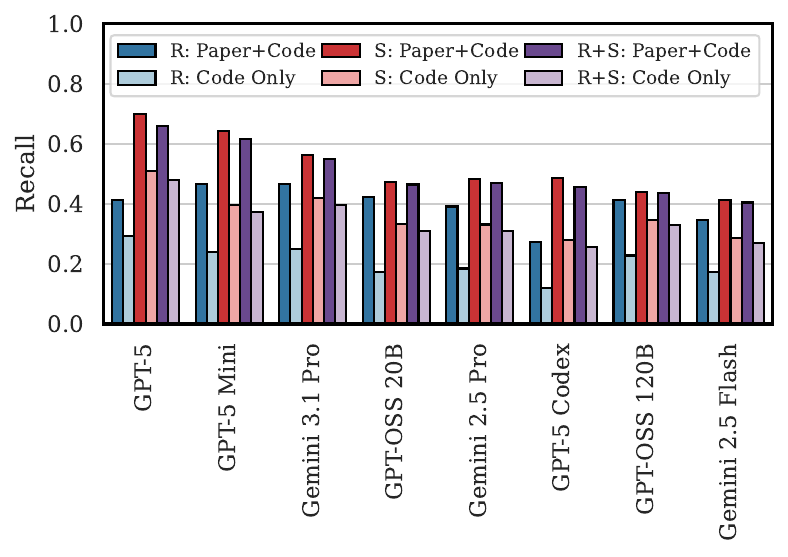}
    \caption{Performance of top 8 models when given paper and code, and only the code, split by data origin: Real (R), Synthetic (S), and combined (R+S).}
    \label{fig:discrepancy-performance-ablation}
\end{figure}

To test the multimodality of our data, we remove the paper from the input, leaving only the code. This makes it more difficult to identify discrepancies, but not impossible, since the repositories also frequently contain abstracts, summaries, or details of the paper in their readme files. Furthermore, scientific papers are typically part of the pre-training data; therefore, while the model does not directly receive the paper as input, it may be able to recall it from its pre-training. We show the results of the top models in this experiment in Fig.~\ref{fig:discrepancy-performance-ablation} (full results in Table~\ref{tab:detailed-results-code-only-ablation}). When removing the paper from the input, all models perform worse, confirming that the paper is necessary to perform the task. We observe an average drop of $19.2$ percentage points, a relative drop of $48.3\%$ on the real data. For the synthetic data, the drop is less pronounced; on average, the models' performance drops by $16.3$ percentage points (relative drop of $30.8\%$).

\subsection{Validation of Unlabeled Discrepancies}\label{sec:validation-unlabeled-discrepancies}
To better understand the precision of the models, we analyze GPT-5's, Gemini 2.5 Pro's, and GPT-OSS 20B's generations. We select predictions from $20$ papers in NLP and CV that were not matched to a ground truth, yielding a total of $224$ evaluations. We select these papers to leverage our own domain expertise, as validating discrepancies is challenging and requires a detailed understanding of the paper and its implementation. We report results of this precision analysis in Table~\ref{tab:precision-analysis}.

Overall, we find that Gemini 2.5 Pro does not make significant errors, achieving a precision of $94.1\%$. Its three errors were due to preprocessing (an omitted term in a formula during OCR), an ambiguity within the paper (a nuance even noted in the model's generation), and the incorrect recall of another paper's formula. GPT-5 follows ($85.7\%$), primarily failing due to paper ambiguities or misinterpretations, or incorrect assumptions about library functions. Finally, GPT-OSS 20B struggles most ($67.0\%$), frequently misunderstanding the code logic (e.g., failing to align variables named differently in the paper and code or misinterpreting conditional execution paths).

Taking inspiration from information retrieval benchmarks where incomplete relevance judgments are common \citep{buckley-voorhees-2004-retrieval-incomplete}, we construct a pooled ground truth from the $20$ papers. We aggregate the discrepancies originally annotated in \scicodeqa with the verified ones identified by the models (excluding minor ones, which are mostly discrepancies related to mismatching settings that can typically be resolved via configuration or the CLI). This results in a comprehensive set of $129$ distinct paper-code discrepancies, and we report the results in Table~\ref{tab:precision-analysis-extended}. We find that the number of discrepancies uniquely identified by the models is high (see also Table~\ref{tab:precision-analysis}), yet recall remains low ($41$--$56\%$), confirming our main experiment's findings that model recall is limited. GPT-5 achieves the best precision-recall tradeoff, obtaining an F1 score of $64.7\%$. We evaluate all $22$ models against this pooled ground truth in \S\ref{sec:pooled-precision-analysis}, confirming that recall remains the primary bottleneck. We make the annotations from the precision analysis public, so future work can leverage them as well for detailed analysis.

\begin{table}[t]
    \small
    \centering
    \begin{tabular}{@{}l*{3}{S[table-format=2.1]}@{}}
    \toprule
    Model & {GPT-5} & {Gemini} & {GPT-OSS} \\
    \midrule
True Positives & 66 & 53 & 72 \\
False Positives & 9 & 3 & 31 \\
\midrule
Precision & 88.0 & 94.6 & 69.9 \\
Recall & 51.2 & 41.1 & 55.8 \\
F1 & 64.7 & 57.3 & 62.1 \\
\bottomrule
    \end{tabular}
    \caption{Performance of GPT-5, Gemini 2.5 Pro, and GPT-OSS 20B on $129$ pooled discrepancies from the annotation of unlabeled predictions and discrepancies from \scicodeqa from $20$ NLP and CV papers.}
    \label{tab:precision-analysis-extended}
\end{table}

\section{Conclusion}
We introduced \scicodeqa, a dataset of $635$ discrepancies designed to evaluate the alignment between a scientific paper and its code. Our analysis reveals a critical gap: while models like GPT-5 and Gemini demonstrate high precision, they suffer from insufficient recall, detecting only $46.7\%$ of real-world discrepancies. For a verification use case, low recall is a particularly consequential failure mode: missed discrepancies silently provide false assurance, while false positives can be filtered by a human reviewer. We further find that models struggle with Paper Omissions, where code logic is absent from the paper text, and fail to maintain performance on recent, uncontaminated publications. Consequently, while current LLMs show promise as assistants, they cannot yet serve as autonomous verifiers of paper-code faithfulness. As research scales toward ``AI Scientists,'' \scicodeqa provides the essential ground truth to ensure these agents remain faithful to the scientific method. Beyond this primary use case, \scicodeqa can serve multiple roles: as a natural long-context reasoning benchmark complementing synthetic retrieval probes, as a testbed for cross-modal claim verification between text and code, and as an evaluation suite for code comprehension in scientific domains. We release the dataset, model generations, evaluation pipeline, and synthetic generation code to support these applications. To further advance this field, we encourage the broader adoption of reproducibility tracks to systematically capture these discrepancies and expand future data collection.

\section{Limitations}\label{sec:limitations}

\paragraph{Domains} Our real-world data is predominantly skewed towards Computer Science and Artificial Intelligence. Applying our real-world pipeline to non-CS domains, searching GitHub repositories for URLs linking to PubMed, bioRxiv, and medRxiv, yielded very few issues reporting paper-code discrepancies, and we did not identify comparable reproducibility tracks outside of CS with paper-code discrepancy reports that we could use. The infrastructure for surfacing paper-code discrepancies thus appears to be largely absent outside CS/AI. Our synthetic pipeline is a short-term response, which we believe provides more value than excluding these domains entirely. Long-term, we call for broader adoption of reproducibility tracks, journals, and projects, such as ReScience C \citep{rougier-etal-2017-rescience} and the new Replication Research journal \citep{roeseler-etal-2025-replication-research}. Consequently, the distribution of errors outside CS/AI may differ from our synthetic approximations, and performance on non-CS domains should be interpreted with this synthetic nature in mind.

\paragraph{Discrepancy Definition} Our discrepancy definition focuses on ``meaningful mismatches'' that impact reproducibility, explicitly excluding simple bugs, hyperparameter configuration mismatches, or documentation nits. While this focuses the task on scientific validity, it means \scicodeqa does not cover the full spectrum of software engineering defects that may exist in research code.

\paragraph{Dataset Size} With $635$ discrepancies, our dataset is relatively small compared to large-scale pre-training corpora. This size is a deliberate trade-off to ensure high quality: real-world discrepancies are naturally sparsely documented, and we employed a rigorous manual verification process to guarantee that every entry constitutes a meaningful mismatch rather than a trivial error or noise.

\paragraph{Synthetic GPT-5 Performance} Our synthetic discrepancies are generated by GPT-5, which is also evaluated on the benchmark. We find that GPT-5 achieves disproportionately higher recall on synthetic data. This suggests that, on this subset, GPT-5 may benefit from its involvement in the data generation. However, the strong correlation between real and synthetic recall across all $22$ models ($r=0.94$, Fig.~\ref{fig:real-vs-synthetic-recall}) confirms that relative model comparisons remain valid. Notably, excluding the GPT-5 family from the correlation analysis increases the correlation to $r=0.98$, confirming that the overall trend is not driven by GPT-5's generation advantage; rather, the GPT-5 family sits slightly off the shared trend line that the other models follow closely. To mitigate this limitation, absolute model performance should be assessed on the real subset, while the synthetic subset reliably supports relative model comparison and allows for targeted generation of new, uncontaminated data.

\section{Ethical Considerations}

\paragraph{Data Release} We constructed the synthetic portion of \scicodeqa using repositories with permissive licenses (MIT, Apache 2.0, BSD, CC-BY) to ensure respectful redistribution of code. For real-world discrepancies derived from GitHub issues and reproducibility reports, we leverage the publicly available data to create our ground truth data by contextualizing and rephrasing the original issues. None of our data contains any personal information; however, since we work with scientific publications, they are closely associated with the authors of the respective papers. We acknowledge that highlighting discrepancies in a specific author's work may be perceived negatively. We emphasize that these discrepancies are treated as scientific artifacts for improving community reproducibility standards, rather than criticisms of individual researchers. Discrepancies frequently arise from benign causes, including concurrent code updates, simplifications for readability, or distinct authorship of paper and code, and their presence does not imply negligence or misconduct.

\paragraph{Automation Risks} The benchmarked models in this work are intended to assist in quality assurance of scientific papers and their codebases. However, there is a risk of over-reliance; given the low recall rates demonstrated in our experiments ($46.7\%$ for Gemini 3.1 Pro and GPT-5 Mini), automated tools should not yet be used as the sole arbiter of a paper's validity. Relying blindly on these systems could lead to a false sense of security regarding the reproducibility and validity of a paper. We emphasize that model outputs should support human review and reproducibility efforts, not serve as automatic reject signals or as evidence of bad faith. Automated detections, particularly given the false positive rates observed in our experiments, require expert verification before any conclusions about research quality are drawn.

\section*{Acknowledgments}
We thank Jan Buchmann, Bhavyajeet Singh, and Vatsal Venkatkrishna for their helpful comments throughout the paper-writing process, as well as the anonymous reviewers and area chair for their constructive feedback.

This work has been funded by the LOEWE Distinguished Chair ``Ubiquitous Knowledge Processing,'' LOEWE initiative, Hesse, Germany (Grant Number: LOEWE/4a//519/05/00.002(0002)/81) and by the German Federal Ministry of Research, Technology and Space and the Hessian Ministry of Higher Education, Research, Science and the Arts within their joint support of the National Research Center for Applied Cybersecurity ATHENE.

\bibliography{bib}

\appendix

\section{Implementation Details}\label{sec:implementation-details}
\subsection{GitHub Crawl Details}\label{sec:github-crawl-details}
We use the GitHub API\footnote{\url{https://docs.github.com/en/rest?apiVersion=2022-11-28}} to search for code repositories that provide code for a research paper. To obtain only repositories for a single paper, we search the description and homepage fields of the repositories, which are commonly used to indicate which paper the code is intended to implement. Specifically, we search for the following URLs: \texttt{arxiv.org}, \texttt{openreview.net}, \texttt{aclanthology.org}, \texttt{doi.org/10.1145}. We further limit our search to only include code repositories that have been published since 2020.

\subsection{Issue Processing}
For the initial GitHub issue classification with Qwen3 4B Thinking, we used a low temperature of $0.2$ to minimize generation variance and promote strict adherence to the classification schema. For the verification of discrepancies from GitHub issues with GPT-5, we checked whether an issue description contained screenshots of the paper or code. If so, we replaced these with their text equivalents.

\subsection{Versioning}
To ensure that the paper and code still contain the discrepancy, we provide links to versioned publications for arXiv, using the version at the time of the reporting of the discrepancy, i.e., publication of the reproducibility report or creation of the GitHub issue. Similarly, for the codebases, we use the commit history to obtain the version of the code at the time of the reporting of the discrepancy. Our dataset contains the versioned links to both the paper (for arXiv) and the codebase. For the synthetic data, we take a version of the codebase as of 31-10-2025.

\begin{table*}[ht]
\centering
\resizebox{\textwidth}{!}{
\begin{tabular}{lllrrr}
\toprule
Model & Paper & Model Card & \makecell{Context\\Window} & \makecell{Knowledge\\Cutoff} & \makecell{Release\\Date} \\ \midrule
DeepSeek Coder 16B v2 & \citetalias{deepseek-coder-v2} (\citeyear{deepseek-coder-v2}) & \href{https://hf.co/deepseek-ai/DeepSeek-Coder-V2-Lite-Instruct}{deepseek-ai/DeepSeek-Coder-V2-Lite-Instruct} & 131k & Nov 2023 & Jun 2024 \\
DeepSeek R1 32B & \citet{deepseek-r1} & \href{https://hf.co/deepseek-ai/DeepSeek-R1-Distill-Qwen-32B}{deepseek-ai/DeepSeek-R1-Distill-Qwen-32B} & 131k & Jul 2024 & Jan 2025 \\
DeepSeek R1 8B & \citet{deepseek-r1} & \href{https://hf.co/deepseek-ai/DeepSeek-R1-Distill-Llama-8B}{deepseek-ai/DeepSeek-R1-Distill-Llama-8B} & 131k & Jul 2024 & Jan 2025 \\
Gemini 2.5 Flash & \citet{gemini2.5}  & \href{https://storage.googleapis.com/deepmind-media/Model-Cards/Gemini-2-5-Flash-Model-Card.pdf}{gemini-2.5-flash} & 1,047k & Jan 2025 & Aug 2025 \\
Gemini 2.5 Flash Lite & \citet{gemini2.5} & \href{https://storage.googleapis.com/deepmind-media/Model-Cards/Gemini-2-5-Flash-Lite-Model-Card.pdf}{gemini-2.5-flash-lite} & 1,047k & Jan 2025 & Aug 2025 \\
Gemini 2.5 Pro & \citet{gemini2.5} & \href{https://storage.googleapis.com/deepmind-media/Model-Cards/Gemini-2-5-Pro-Model-Card.pdf}{gemini-2.5-pro} & 1,047k & Jan 2025 & Aug 2025 \\
Gemini 3.1 Pro & \citet{gemini3pro2025} & \href{https://storage.googleapis.com/deepmind-media/Model-Cards/Gemini-3-1-Pro-Model-Card.pdf}{gemini-3.1-pro-preview} & 1,047k & Jan 2025 & Feb 2026 \\
GPT-OSS 120B & \citet{gpt-oss-koch-et-al-2025} & \href{https://hf.co/openai/gpt-oss-120b}{openai/gpt-oss-120b} & 131k & Jun 2024 & Aug 2025 \\
GPT-OSS 20B & \citet{gpt-oss-koch-et-al-2025} & \href{https://hf.co/openai/gpt-oss-20b}{openai/gpt-oss-20b} & 131k & Jun 2024 & Aug 2025 \\
GPT-5 & \citet{gpt-5} & \href{https://platform.openai.com/docs/models/gpt-5}{gpt-5-2025-08-07} & 272k & Sep 2024 & Aug 2025 \\
GPT-5 Codex & \citet{gpt-5} & \href{https://platform.openai.com/docs/models/gpt-5-codex}{gpt-5-codex} & 272k & Sep 2024 & Aug 2025 \\
GPT-5 Mini & \citet{gpt-5} & \href{https://platform.openai.com/docs/models/gpt-5-mini}{gpt-5-mini-2025-08-07} & 272k & May 2024 & Aug 2025 \\
GPT-5 Nano & \citet{gpt-5} & \href{https://platform.openai.com/docs/models/gpt-5-nano}{gpt-5-nano-2025-08-07} & 272k & May 2024 & Aug 2025 \\
Devstral 24B Small & \citet{devstral} & \href{https://hf.co/mistralai/Devstral-Small-2507}{mistralai/Devstral-Small-2507} & 131k & Mar 2025 & Jul 2025 \\
Magistral 24B Small & \citet{magistral} & \href{https://hf.co/mistralai/Magistral-Small-2509}{mistralai/Magistral-Small-2509} & 131k & Jun 2025 & Sep 2025 \\
Nemotron Nano 9B v2 & \citet{nemotron-nano} & \href{https://hf.co/nvidia/NVIDIA-Nemotron-Nano-9B-v2}{nvidia/NVIDIA-Nemotron-Nano-9B-v2} & 131k & Apr 2025 & Aug 2025 \\
Nemotron Super 49B v1.5 & \citet{llama-nemotron} & \href{https://hf.co/nvidia/Llama-3_3-Nemotron-Super-49B-v1_5}{nvidia/Llama-3\_3-Nemotron-Super-49B-v1\_5} & 131k & Dec 2023 & Jul 2025 \\
Qwen3 30B Coder & \citet{yang-etal-qwen3-2025} & \href{https://hf.co/Qwen/Qwen3-Coder-30B-A3B-Instruct}{Qwen/Qwen3-Coder-30B-A3B-Instruct} & 262k & Mar 2025 & Jul 2025 \\
Qwen3 30B Instruct & \citet{yang-etal-qwen3-2025} & \href{https://hf.co/Qwen/Qwen3-30B-A3B-Instruct-2507}{Qwen/Qwen3-30B-A3B-Instruct-2507} & 262k & Mar 2025 & Jul 2025 \\
Qwen3 30B Thinking & \citet{yang-etal-qwen3-2025} & \href{https://hf.co/Qwen/Qwen3-30B-A3B-Thinking-2507}{Qwen/Qwen3-30B-A3B-Thinking-2507} & 262k & Mar 2025 & Jul 2025 \\
Qwen3 4B Instruct & \citet{yang-etal-qwen3-2025} & \href{https://hf.co/Qwen/Qwen3-4B-Instruct-2507}{Qwen/Qwen3-4B-Instruct-2507} & 262k & Mar 2025 & Jul 2025 \\
Qwen3 4B Thinking & \citet{yang-etal-qwen3-2025} & \href{https://hf.co/Qwen/Qwen3-4B-Thinking-2507}{Qwen/Qwen3-4B-Thinking-2507} & 262k & Mar 2025 & Jul 2025 \\ 
\bottomrule
\end{tabular}
}
\caption{Model versions including their context window, knowledge cutoff (date of most recent pre-training data), and release date.}
\label{tab:model-version}
\end{table*}
\subsection{Paper Processing}
We provide the paper text in markdown as input. We use Mistral's OCR \citep{mistral-2025-ocr} to convert the PDF to markdown. We exclude figures, although the captions are included. We further remove the references section from all papers.

\subsection{Code Processing}
To provide the code in the prompt, we obtain all files from the GitHub repository at a target date using the commit history, which we set to the date when the original GitHub issue was raised or the reproducibility paper was published. For the synthetic data, we set the cutoff to 31-10-2025. Having a versioned and static code repository is crucial to ensure reproducibility of our experiments and to make sure paper-code discrepancies have not been fixed in our dataset.

We retain only files that contain relevant content, mainly excluding dataset files. Specifically, we exclude files that are larger than 1MB and include files with the following extensions: \texttt{.c, .cc, .cpp, .cu, .h, .hpp, .java, .jl, .m, .matlab, Makefile, .md, .pl, .ps1, .py, .r, .sh, config.txt, .rs, readme.txt, requirements\_dev.txt, requirements-dev.txt, requirements.dev.txt, requirements.txt, .scala, .yaml, .yml}. We obtain this list by manually inspecting the file extensions in the repositories of our dataset. We further convert all Jupyter notebooks to Python files by removing their outputs. For the prompt, we further include a list of all files in the repository.

\subsection{Context Processing}
The markdown paper and processed code repository are inserted into the prompt. We fill the prompt up to $90\%$ of the context size to leave space for the reasoning (if supported by the model) and output. If the paper and code would exceed the context window, we truncate entire files from the end of the code prompt until it fits into the context window.

\subsection{Model and Decoding Configuration}
We deploy the open-weight models with FP16, except for the GPT-OSS models, which were only published in MXFP4. We utilize the Ollama and vLLM libraries \citep{vllm} on 1-4 A100 80GB GPUs, depending on the model's VRAM requirements; the commercial models are queried via their respective APIs. We generally set the temperature to $1.0$, and for the GPT-5, GPT-OSS, and Gemini models, set a high reasoning effort or budget to maximize performance. For the Gemini models, we set a high reasoning budget of $24k$ tokens; for the GPT-5 models and GPT-OSS, we set a ``high'' reasoning budget. Further details about each model can be obtained from the respective model card linked in Table~\ref{tab:model-version}.

\begin{figure*}[htpb]
    \centering
    \includegraphics[width=1\linewidth]{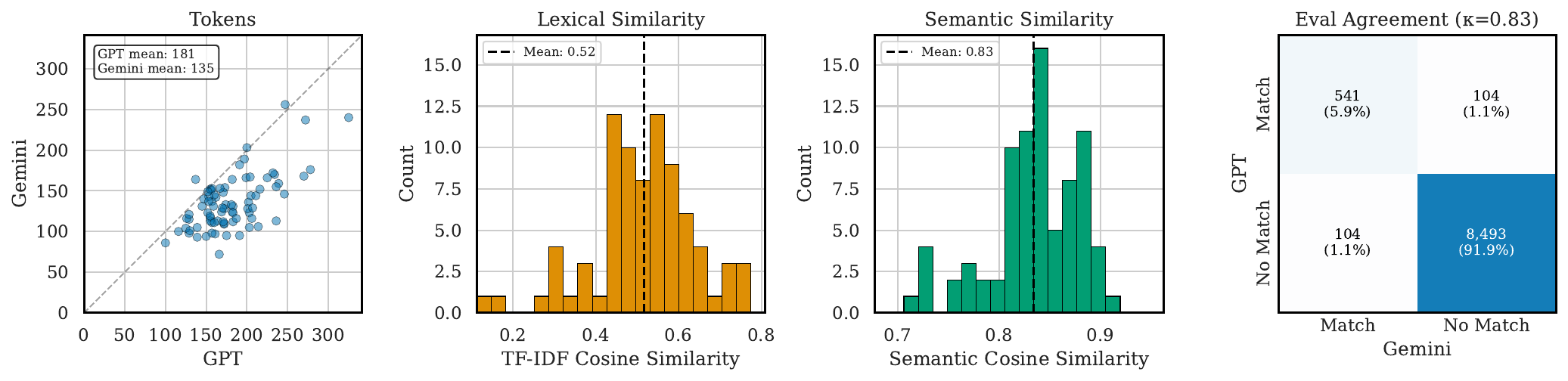}
    \caption{Comparisons between the two ground truth descriptions generated by Gemini 3.1 Pro and GPT-5.}
    \label{fig:dual-gt-analysis}
\end{figure*}

\begin{figure*}[htpb]
    \centering
    \includegraphics[width=1\linewidth]{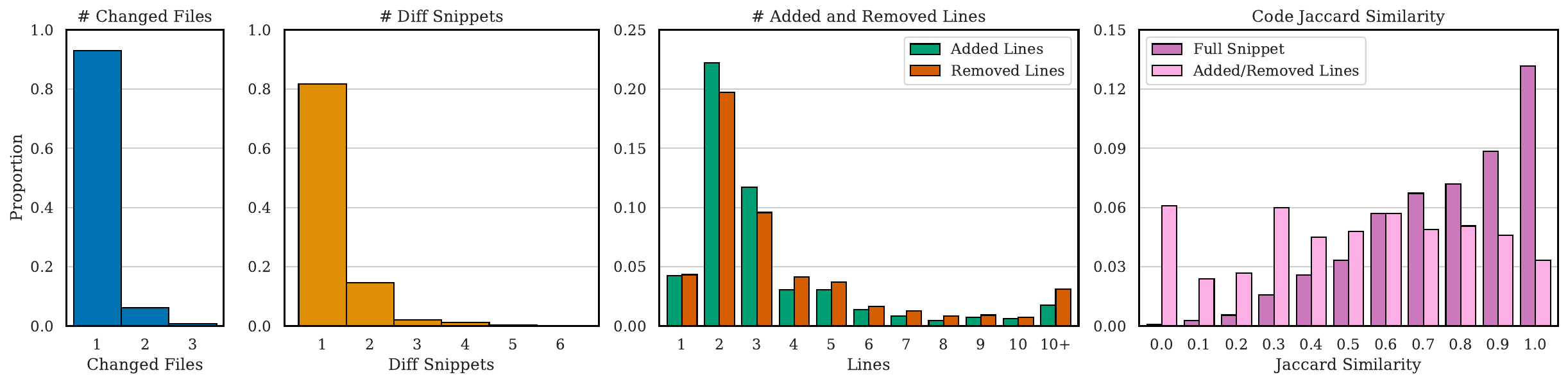}
    \caption{Quantitative analysis of synthetic code modifications. We show the distribution of number of changed files, number of generated diff snippets, added (green) vs. removed (orange) lines in those snippets, and Jaccard similarity between the original and modified code.}
    \label{fig:synthetic-code-analysis}
\end{figure*}

\section{Extended Analysis}
\subsection{Gemini and GPT Ground Truth Comparison}
For the real data, we rephrased the ground truth discrepancy descriptions with Gemini 3.1 Pro and GPT-5 (see \S\ref{sec:data-collection}). We analyze the similarity between these two descriptions in Fig.~\ref{fig:dual-gt-analysis}. We find the GPT descriptions to be slightly longer ($181\pm40$ vs. $135\pm34$ tokens). We further find moderate lexical overlap between the two ground truths (average TF-IDF cosine similarity is $0.52$), but high semantic similarity (average $0.83$ computed using EmbeddingGemma), indicating that the models use different wording to describe the same discrepancy. Finally, we analyze how often the GPT-OSS 20B judge matches a prediction against none, either, or both ground truths. We find that in $97.7\%$ of cases, the judge comes to the same verdict given the two different discrepancy descriptions, yielding a Cohen's Kappa of $\kappa=0.83$.

\subsection{Synthetic Code Analysis}\label{sec:analysis-synthetic-code}
We conduct a quantitative analysis of the synthetic data to validate that our generation pipeline produces minimal changes, leaving the main code intact. Fig.~\ref{fig:synthetic-code-analysis} presents the analysis of the generated diffs.

\paragraph{Code Changes} We find the generated discrepancies in the code to be highly targeted. The vast majority affect only a single file ($1.08\pm0.29$ on average), with a maximum of three files. Similarly, the number of distinct generated diffs is low, with a mean of $1.24\pm0.61$ snippets per discrepancy. This confirms that the generated errors are specific deviations in logic rather than large-scale refactors. The number of line edits (i.e., added or removed lines) further supports the subtlety of the dataset. The distribution of line counts is heavily skewed towards small edits: On average, $2.41\pm3.25$ lines are added and $3.13\pm4.47$ lines are removed. The slightly higher count for removed lines suggests the discrepancies often simplify logic (e.g., removing a normalization step) rather than adding complex boilerplate.

\paragraph{Code Similarity} To quantify the similarity between the original and modified code, we calculate the Jaccard similarity.
We measure the similarity at two levels: over the entire generated code block (which often contains a few unchanged lines) and over only the code lines that were modified. Comparing the entire code block generated by the model against the original yields a high mean similarity of $0.74\pm0.20$. This indicates that the surrounding context and structure remain largely identical. Comparing only the specific lines that differ (added vs. removed) yields a mean similarity of $0.50\pm0.29$. This moderate overlap suggests the changes are often modifications of existing statements (e.g., changing an operator or variable) rather than complete replacements. Overall, the data confirms that the synthetic discrepancies are precise and minimal, strictly adhering to the ``small'' and ``conceptually meaningful'' constraints of the synthetic data generation prompt (\S\ref{sec:prompts-synthetic-discrepancy-generation}).

\subsection{Programming Language Distribution}\label{sec:analysis-programming-language}
\begin{figure*}[tpb]
    \centering
    \includegraphics[width=\linewidth]{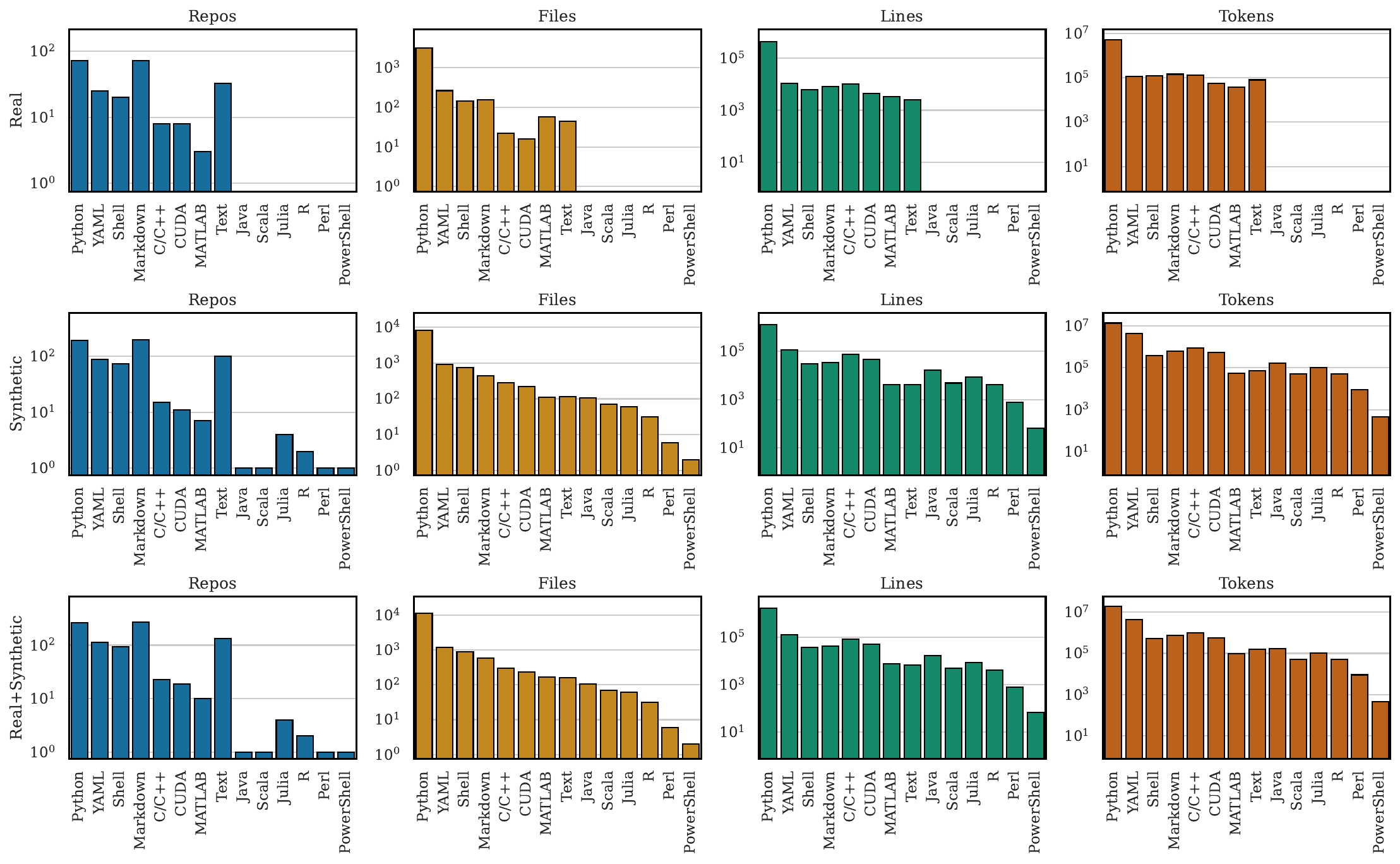}
    
    \caption{Distribution of programming languages in \scicodeqa. The columns of the plot show the distribution by number of \textit{Repos}, \textit{Files}, \textit{Lines}, and \textit{Tokens} for each respective programming language. The rows show the distribution for the real, synthetic, and combined data.}
    \label{fig:programming-language-distribution}
\end{figure*}

Fig.~\ref{fig:programming-language-distribution} shows the distribution of programming languages in the \scicodeqa dataset. Since the real data originates from the CS/AI domain, we observe Python as the dominant language, along with a small portion of C/C++, MATLAB, and CUDA files. Besides Python, other languages are present in the synthetic data, including Java, Scala, Julia, and R, although their overall presence is relatively small.

\section{Extended Results}
\subsection{Error Analysis}
\begin{table}[tpb]
    \small
    \centering
    \resizebox{\linewidth}{!}{
    \begin{tabular}{@{}l*{3}{S[table-format=2.1]}@{}}
    \toprule
    Model & {GPT-5} & {Gemini} & {GPT-OSS} \\
    \midrule
    False Positives & 10 & 3 & 34 \\
    \hspace{3mm}\textit{Code Misunderstanding} & 1 & 0 & 20 \\
    \hspace{3mm}\textit{Paper Misunderstanding} & 2 & 0 & 9 \\
    \hspace{3mm}\textit{Paper Ambiguity} & 2 & 1 & 0 \\
    \hspace{3mm}\textit{OCR Error} & 1 & 1 & 1 \\
    \hspace{3mm}\textit{3\textsuperscript{rd} Party Code} & 2 & 0 & 1 \\
    \hspace{3mm}\textit{No Discrepancy} & 1 & 0 & 1 \\
    \hspace{3mm}\textit{Minor (Config)} & 1 & 0 & 2 \\
    \hspace{3mm}\textit{Other} & 0 & 1 & 0 \\
    \midrule
    True Positives & 60 & 48 & 69 \\
    \hspace{3mm}\textit{Unique} & 23 & 19 & 34 \\
    \hspace{3mm}\textit{Minor} & 7 & 7 & 12 \\
    Total & 70 & 51 & 103 \\
    Precision & 85.7 & 94.1 & 67.0 \\
    \bottomrule
    \end{tabular}
    }
    \caption{Analysis of unlabeled discrepancy predictions of GPT-5, Gemini 2.5 Pro, and GPT-OSS 20B on $20$ NLP and CV papers from the real subset of \scicodeqa.
    }
    \label{tab:precision-analysis}
\end{table}

Table~\ref{tab:precision-analysis} provides the detailed breakdown of our human analysis from \S\ref{sec:validation-unlabeled-discrepancies} on the predictions of GPT-5, Gemini 2.5 Pro, and GPT-OSS 20B.

\subsection{Performance by Programming Language}
In the synthetic dataset, the known injection points allow us to analyze detection performance across different programming languages. Since the dataset is skewed towards Python, drawing conclusions for specific low-resource languages is difficult due to the small sample size. Therefore, we group all non-Python instances into a unified subset of $46$ samples for analysis (``Not Python''). The resulting recall scores are presented in Table~\ref{tab:performance-by-language}. Most models show only small, inconsistent differences between Python and non-Python recall (within $\pm 5$\,pp). The notable exceptions are Devstral 24B Small, which collapses from $15.1\%$ on Python to $0\%$ on non-Python, and Gemini 2.5 Flash ($+9.7$\,pp); conversely, GPT-5 Mini ($-5.9$\,pp), GPT-OSS 20B ($-5.4$\,pp), and Nemotron Nano 9B v2 ($-7.8$\,pp) actually perform slightly better on non-Python. Beyond these aggregate trends, two specific languages stand out. MATLAB is a positive outlier: GPT-5 and GPT-5 Mini both achieve $90.0\%$ recall, substantially above their Python performance. With $n=10$ this is tentative, but we hypothesize that MATLAB's mathematical syntax aligns more closely with the equations presented in scientific papers, facilitating easier logical alignment for the models. Conversely, CUDA is a near-universal weak spot: $17$ of $22$ models score $0\%$ recall on CUDA ($n=4$), including GPT-5, Gemini 3.1 Pro, and GPT-OSS 120B. Small sample sizes limit definitive conclusions for both languages.

\begin{table}[tpb]
\centering
\begin{tabular}{ll}
\toprule
\makecell[l]{Code File\\Prompt Position} & \# Samples \\
\midrule
$\leq32k$ &  $253$ ($39.8\%$) \\
$\leq64k$ &  $463$ ($72.9\%$) \\
$\leq131k$ & $568$ ($89.4\%$) \\
$\leq262k$ & $631$ ($99.4\%$) \\
$\leq524k$ & $635$ ($100.0\%$) \\
\bottomrule
\end{tabular}
\caption{Number of data samples by position of the last relevant code file in the prompt.}
\label{tab:truncation-analysis}
\end{table}
\subsection{Truncation Analysis}\label{sec:truncation-analysis}

\begin{table*}[tpb]
\resizebox{\textwidth}{!}{
\begin{tabular}{lS[table-format=2.1]S[table-format=2.1]S[table-format=2.1]S[table-format=2.1]S[table-format=2.1]S[table-format=2.1]S[table-format=2.1]S[table-format=2.1]S[table-format=2.1]S[table-format=2.1]S[table-format=2.1]}
\toprule
{Model} & {Python} & {Not Python} & {Julia} & {MATLAB} & {YAML} & {C/C++} & {CUDA} & {R} & {Shell} & {Py+YAML} & {AVG} \\
\midrule
\multicolumn{1}{r}{\# Discrepancies} & {496} & {46} & {12} & {10} & {9} & {5} & {4} & {3} & {3} & {1} & {543} \\
\midrule
GPT-5 & 70.4 & 65.2 & 66.7 & 90.0 & 75.0 & 60.0 & 0.0 & 100.0 & 33.3 & 100.0 & 70.0 \\
GPT-5 Mini & 63.7 & 69.6 & 75.0 & 90.0 & 66.7 & 60.0 & 25.0 & 100.0 & 33.3 & 100.0 & 64.3 \\
Gemini 3.1 Pro & 56.9 & 50.0 & 50.0 & 80.0 & 50.0 & 60.0 & 0.0 & 66.7 & 0.0 & 100.0 & 56.4 \\
GPT-OSS 20B & 46.8 & 52.2 & 50.0 & 60.0 & 66.7 & 60.0 & 50.0 & 33.3 & 0.0 & 0.0 & 47.1 \\
Gemini 2.5 Pro & 48.8 & 45.7 & 41.7 & 60.0 & 50.0 & 40.0 & 50.0 & 33.3 & 33.3 & 0.0 & 48.4 \\
GPT-5 Codex & 48.8 & 45.7 & 33.3 & 70.0 & 50.0 & 40.0 & 25.0 & 66.7 & 33.3 & 100.0 & 48.6 \\
GPT-OSS 120B & 44.6 & 39.1 & 33.3 & 80.0 & 41.7 & 40.0 & 0.0 & 33.3 & 0.0 & 0.0 & 44.0 \\
Gemini 2.5 Flash & 42.3 & 32.6 & 33.3 & 40.0 & 33.3 & 40.0 & 0.0 & 66.7 & 33.3 & 0.0 & 41.4 \\
GPT-5 Nano & 27.2 & 30.4 & 33.3 & 30.0 & 41.7 & 40.0 & 0.0 & 33.3 & 0.0 & 0.0 & 27.4 \\
Nemotron Super 49B v1.5 & 24.0 & 23.9 & 16.7 & 20.0 & 16.7 & 40.0 & 50.0 & 33.3 & 33.3 & 0.0 & 23.9 \\
Qwen3 30B Coder & 24.0 & 21.7 & 25.0 & 30.0 & 41.7 & 0.0 & 0.0 & 33.3 & 0.0 & 0.0 & 23.8 \\
Gemini 2.5 Flash Lite & 24.0 & 19.6 & 16.7 & 20.0 & 50.0 & 20.0 & 0.0 & 0.0 & 0.0 & 100.0 & 23.8 \\
Nemotron Nano 9B v2 & 16.1 & 23.9 & 25.0 & 40.0 & 50.0 & 0.0 & 0.0 & 0.0 & 0.0 & 0.0 & 16.8 \\
Qwen3 30B Inst. & 22.0 & 19.6 & 25.0 & 20.0 & 33.3 & 0.0 & 0.0 & 33.3 & 0.0 & 0.0 & 21.7 \\
DeepSeek R1 32B & 16.7 & 13.0 & 8.3 & 20.0 & 8.3 & 0.0 & 0.0 & 66.7 & 0.0 & 0.0 & 16.4 \\
DeepSeek Coder 16B V2 & 10.1 & 4.3 & 8.3 & 10.0 & 0.0 & 0.0 & 0.0 & 0.0 & 0.0 & 0.0 & 9.6 \\
Qwen3 4B Inst. & 16.7 & 8.7 & 16.7 & 10.0 & 0.0 & 0.0 & 0.0 & 33.3 & 0.0 & 0.0 & 16.0 \\
Magistral 24B Small & 14.5 & 10.9 & 25.0 & 0.0 & 25.0 & 0.0 & 0.0 & 0.0 & 0.0 & 0.0 & 14.2 \\
Devstral 24B Small & 15.1 & 0.0 & 0.0 & 0.0 & 0.0 & 0.0 & 0.0 & 0.0 & 0.0 & 100.0 & 14.0 \\
Qwen3 30B Think. & 15.5 & 15.2 & 8.3 & 30.0 & 8.3 & 20.0 & 0.0 & 33.3 & 0.0 & 0.0 & 15.5 \\
Qwen3 4B Think. & 11.3 & 8.7 & 8.3 & 0.0 & 41.7 & 0.0 & 0.0 & 0.0 & 0.0 & 0.0 & 11.0 \\
DeepSeek R1 8B & 5.6 & 0.0 & 0.0 & 0.0 & 0.0 & 0.0 & 0.0 & 0.0 & 0.0 & 0.0 & 5.2 \\
\bottomrule
\end{tabular}
}
\caption{Discrepancy recall by programming language on the synthetic \scicodeqa data.}
\label{tab:performance-by-language}
\end{table*}

In \S\ref{sec:results} we analyzed the model performance by input length, finding that longer inputs degrade performance. We extend this analysis by investigating the performance as a function of the position of the last relevant code file in the prompt. The relevant code files are obtained in the validation/rephrasing step for the real data, and during discrepancy generation for the synthetic data. Table~\ref{tab:truncation-analysis} shows the distribution of the position of the last relevant code file in the prompt. We find that $89.4\%$ of samples in our data contain all relevant code files within a context window of $131k$ tokens, and $99.4\%$ within a $262k$ context window. 

We further analyze recall by the position of the last relevant code file in the prompt, binning the samples and reporting results in Table~\ref{tab:truncation-analysis-performance}. We observe substantial recall degradation even for models that experience virtually zero truncation. GPT-5 drops from $77.5\%$ to $50.7\%$ and Gemini 3.1 Pro from $64.4\%$ to $37.3\%$ as the relevant file appears later in the prompt. 

Both the code file position distribution and the performance degradation demonstrate that the dominant factor is long-context reasoning, not truncation, consistent with our analysis in Fig.~\ref{fig:discrepancy-results} and prior findings on long-context degradation \citep{liu-etal-2024-lost, ruler-hsieh-etal-2024,levy-etal-2024-task}.

\begin{figure}[t]
    \centering
    \includegraphics[width=1\linewidth]{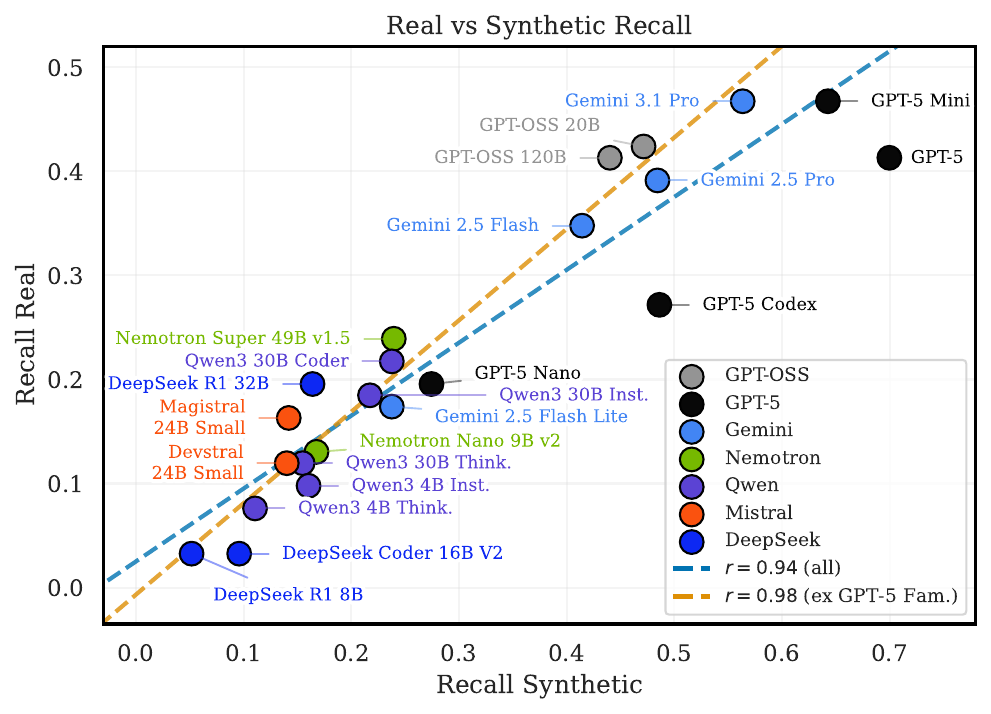}
    \caption{Correlation between model recall on the synthetic (x-axis) and real (y-axis)  subsets of \scicodeqa. Each point represents one of the $22$ evaluated models. The dashed lines show linear fits across all models (blue) or excluding the GPT-5 family (orange). The $r$ values denote Pearson correlation coefficients.}
    \label{fig:real-vs-synthetic-recall}
\end{figure}

\subsection{Real vs. Synthetic Recall}\label{sec:real-vs-synthetic-recall}

\begin{table}[t]
\centering
\resizebox{\linewidth}{!}{
\begin{tabular}{@{}l*{4}{S[table-format=3.1, table-column-width=1.1cm]}@{}}
\toprule
\makecell[lb]{\\Model} & {{\makecell[b]{Gemini\\+GPT}}} & {{\makecell[b]{Gemini}}}& {{\makecell[b]{GPT}}} & {{\makecell[b]{$\Delta$}}} \\
\midrule
GPT-5 & 45.0 & 45.0 & 43.8 & -1.3 \\
GPT-5 Mini & 48.8 & 47.5 & 43.8 & -3.7 \\
Gemini 3.1 Pro & 50.0 & 50.0 & 48.8 & -1.3 \\
GPT-OSS 20B & 47.5 & 45.0 & 42.5 & -2.5 \\
Gemini 2.5 Pro & 41.2 & 40.0 & 40.0 & 0.0 \\
GPT-5 Codex & 27.5 & 27.5 & 27.5 & 0.0 \\
GPT-OSS 120B & 42.5 & 37.5 & 40.0 & 2.5 \\
Gemini 2.5 Flash & 36.2 & 32.5 & 33.8 & 1.3 \\
GPT-5 Nano & 21.2 & 18.8 & 18.8 & 0.0 \\
Nemotron 49B v1.5 & 26.2 & 23.8 & 21.2 & -2.5 \\
Qwen3 30B Coder & 23.8 & 17.5 & 20.0 & 2.5 \\
Gemini 2.5 Flash Lite & 18.8 & 16.2 & 16.2 & 0.0 \\
Nemotron Nano 9B v2 & 13.8 & 8.8 & 11.2 & 2.5 \\
Qwen3 30B Inst. & 20.0 & 13.8 & 17.5 & 3.7 \\
DeepSeek R1 32B & 21.2 & 17.5 & 17.5 & 0.0 \\
DeepSeek Coder 16B V2 & 2.5 & 2.5 & 1.2 & -1.2 \\
Qwen3 4B Inst. & 10.0 & 7.5 & 8.8 & 1.2 \\
Magistral 24B Small & 16.2 & 15.0 & 13.8 & -1.2 \\
Devstral 24B Small & 12.5 & 8.8 & 11.2 & 2.5 \\
Qwen3 30B Think. & 13.8 & 12.5 & 13.8 & 1.3 \\
Qwen3 4B Think. & 8.8 & 7.5 & 8.8 & 1.2 \\
DeepSeek R1 8B & 3.8 & 3.8 & 2.5 & -1.2 \\
\bottomrule
\end{tabular}
}
\caption{Model performance on the real-world \scicodeqa data for the $80$ instances where both a GPT-5 and Gemini 3.1 Pro ground truth discrepancy description is available. The columns show the recall performance when assessed against both (Gemini+GPT), Gemini-only, or GPT-only ground truth. A negative $\Delta$ shows a preference for Gemini's ground truth descriptions, while a positive $\Delta$ shows a preference for GPT's. }
\label{tab:gt-source-performance}
\end{table}
To assess the validity of our synthetic data generation pipeline, we analyze the relationship between model performance on real-world discrepancies versus synthetic discrepancies. Fig.~\ref{fig:real-vs-synthetic-recall} visualizes the recall and the relationship between the subsets. We observe a high positive Pearson correlation of $r=0.94$. When excluding the GPT-5 family (i.e., GPT-5, GPT-5 Mini, GPT-5 Nano, and GPT-5 Codex) from the correlation analysis, the Pearson correlation rises to $r=0.98$, indicating that the GPT-5 family sits slightly off the shared trend line that the other models follow closely. Apart from this caveat, the relative ranking of models remains consistent regardless of the data origin; models that demonstrate strong capabilities on our synthetic injections are reliably better at identifying real-world discrepancies. While this trend might be influenced by the general capabilities of the models, the strong correlation confirms that the synthetic data preserves relative model rankings. Consequently, absolute performance should be assessed on the real subset, while the synthetic subset serves as a reliable proxy for relative model comparison, justifying its use for scaling the benchmark to scientific domains where real-world examples are scarce.

\subsection{Gemini vs. GPT Ground Truth Recall}\label{sec:gemini-vs-gpt-ground-truth}

A potential concern with the LLM-rephrased ground truth discrepancy descriptions is that a model from the same family might have an advantage when evaluated against the ground truth (GT) that was generated by the same model or from the same family. To test this, we compare per-model recall when evaluating against only the GPT-generated versus Gemini-generated ground truth descriptions across the $80$ discrepancies for which both are available. We report results in Table~\ref{tab:gt-source-performance}. We find no evidence of systematic provider bias. In fact, the closed OpenAI models perform the same or better against the Gemini-generated ground truth. Gemini 3.1 Pro performs slightly better against its own generations ($\Delta=-1.3$), but the other Gemini models do not exhibit any bias towards these descriptions. Among third-party model families, which serve as unbiased controls since they are not affiliated with either GT provider, results are similarly balanced: DeepSeek and NVIDIA models show no consistent preference, while the Qwen models slightly favor GPT GT (mean $\Delta = +2.0$pp). Across all $22$ models, the overall mean $\Delta$ is $+0.2$pp with a maximum absolute difference of just $3.7$pp, and $9$ models favor GPT GT versus $8$ favoring Gemini GT ($5$ tied). These small, inconsistent differences suggest that the choice of ground truth source does not introduce meaningful evaluation bias.

\subsection{Pooled Ground Truth Results}\label{sec:pooled-precision-analysis}
\begin{table}[t]
\centering
\begin{tabular}{@{}l*{3}{S[table-format=2.1]}@{}}
\toprule
Model & {P} & {R} & {F1} \\
\midrule
GPT-5 & 84.1 & 63.6 & 72.4 \\
GPT-5 Mini & 70.7 & 62.8 & 66.5 \\
GPT-OSS 20B & 68.9 & 63.6 & 66.1 \\
Gemini 2.5 Pro & 85.7 & 48.8 & 62.2 \\
Gemini 3.1 Pro & 76.1 & 50.4 & 60.6 \\
GPT-OSS 120B & 61.1 & 52.7 & 56.6 \\
Gemini 2.5 Flash & 63.2 & 41.1 & 49.8 \\
GPT-5 Codex & 79.1 & 34.9 & 48.4 \\
Gemini 2.5 Flash Lite & 57.1 & 30.2 & 39.5 \\
Nemotron Super 49B v1.5 & 56.1 & 30.2 & 39.3 \\
GPT-5 Nano & 44.3 & 34.1 & 38.5 \\
Devstral 24B Small & 45.6 & 27.9 & 34.6 \\
Qwen3 30B Inst. & 42.6 & 27.1 & 33.2 \\
Qwen3 30B Coder & 39.1 & 27.9 & 32.6 \\
Qwen3 30B Think. & 62.5 & 18.6 & 28.7 \\
Qwen3 4B Inst. & 38.9 & 22.5 & 28.5 \\
DeepSeek R1 32B & 32.8 & 22.5 & 26.7 \\
Magistral 24B Small & 28.7 & 24.0 & 26.2 \\
Qwen3 4B Think. & 61.5 & 14.7 & 23.8 \\
Nemotron Nano 9B v2 & 23.5 & 17.8 & 20.3 \\
DeepSeek Coder 16B V2 & 11.1 & 6.2 & 8.0 \\
DeepSeek R1 8B & 14.3 & 1.6 & 2.8 \\
\bottomrule
\end{tabular}
\caption{Precision (P), Recall (R), and F1 scores on $20$ NLP and CV papers with $129$ discrepancies. Ground truths have been pooled from the predictions of GPT-5, Gemini 2.5 Pro, and GPT-OSS 20B. Additionally, discrepancies from the real-world data of \scicodeqa have been added.}
\label{tab:pooled-precision-analysis}
\end{table}

In \S\ref{sec:validation-unlabeled-discrepancies} we created a pooled ground truth for $20$ NLP and CV papers by manually verifying the predicted discrepancies from GPT-5, Gemini 2.5 Pro, and GPT-OSS 20B. Together with the existing discrepancies from the real-world data in \scicodeqa on these papers, this yields $129$ verified paper-code discrepancies. Following the pooling methodology from information retrieval \citep{buckley-voorhees-2004-retrieval-incomplete}, where relevance judgments are aggregated from multiple systems to approximate a more complete ground truth, we evaluate all models against the pooled set. In case multiple ground truths exist for the same discrepancy, e.g., when both GPT-5 and Gemini 2.5 Pro surfaced the same unannotated discrepancy, we consider it detected if the prediction matches at least one description. Table~\ref{tab:pooled-precision-analysis} reports the results.

GPT-5 achieves the highest F1 ($72.4\%$), balancing precision ($84.1\%$) and recall ($63.6\%$) the best, followed by GPT-5 Mini and GPT-OSS 20B ($66.5\%$ and $66.1\%$). As established in the pooling literature, systems contributing to the pool have a structural advantage, so the evaluations of GPT-5, Gemini 2.5 Pro, and GPT-OSS 20B should be viewed as upper bounds \citep{buckley-voorhees-2004-retrieval-incomplete}. Nonetheless, clear patterns emerge: a precision--recall tradeoff is evident across models, with Gemini 2.5 Pro reaching the highest precision ($85.7\%$) at the cost of recall ($48.8\%$). GPT-OSS 20B and 120B are the strongest open-weight models, while the remaining open-weight models lag behind considerably (next best: Nemotron 49B, F1 of $39.3\%$). Despite the pool bias favoring contributing systems, recall remains the primary bottleneck: even under favorable conditions, the best model detects at most $63.6\%$ of verified discrepancies.

\begin{table*}[tb]
\centering
\begin{tabular}{@{}l*{5}{S[table-format=3.1, table-column-width=1.1cm]}@{}}
\toprule
 \hfill Bin & All & {$0$-$32k$} & {$32$-$64k$} & {$64$-$131k$} & {$\ge131k$} \\
Model \hfill \# Samples & {$635$} & {$253$} & {$210$} & {$105$} & {$67$} \\
\midrule
GPT-5 & 65.8 & 77.5 & 62.9 & 53.3 & 50.7 \\
GPT-5 Mini & 61.7 & 75.1 & 59.5 & 44.8 & 44.8 \\
Gemini 3.1 Pro & 55.0 & 64.4 & 55.7 & 41.9 & 37.3 \\
GPT-OSS 20B & 46.5 & 60.1 & 41.9 & 35.2 & 26.9 \\
Gemini 2.5 Pro & 47.1 & 58.1 & 43.8 & 37.1 & 31.3 \\
GPT-5 Codex & 45.5 & 57.7 & 41.9 & 36.2 & 25.4 \\
GPT-OSS 120B & 43.6 & 59.3 & 38.6 & 28.6 & 23.9 \\
Gemini 2.5 Flash & 40.5 & 51.4 & 40.5 & 24.8 & 23.9 \\
GPT-5 Nano & 26.3 & 34.8 & 22.4 & 21.9 & 13.4 \\
Nemotron Super 49B v1.5 & 23.9 & 29.6 & 20.5 & 20.0 & 19.4 \\
Qwen3 30B Coder & 23.5 & 27.3 & 20.0 & 24.8 & 17.9 \\
Gemini 2.5 Flash Lite & 22.8 & 33.6 & 18.1 & 14.3 & 10.4 \\
Nemotron Nano 9B v2 & 16.2 & 21.7 & 13.8 & 14.3 & 6.0 \\
Qwen3 30B Inst. & 21.3 & 26.5 & 20.5 & 12.4 & 17.9 \\
DeepSeek R1 32B & 16.9 & 25.7 & 16.7 & 4.8 & 3.0 \\
DeepSeek Coder 16B V2 & 8.7 & 13.8 & 4.3 & 5.7 & 7.5 \\
Qwen3 4B Inst. & 15.1 & 21.7 & 12.9 & 8.6 & 7.5 \\
Magistral 24B Small & 14.5 & 20.9 & 14.3 & 6.7 & 3.0 \\
Devstral 24B Small & 13.7 & 19.0 & 11.9 & 5.7 & 11.9 \\
Qwen3 30B Think. & 15.0 & 19.4 & 14.3 & 7.6 & 11.9 \\
Qwen3 4B Think. & 10.6 & 15.4 & 11.4 & 3.8 & 0.0 \\
DeepSeek R1 8B & 4.9 & 5.5 & 4.8 & 2.9 & 6.0 \\
\bottomrule
\end{tabular}
\caption{Model performance by position of the last relevant code file in the prompt. The \textit{All} column considers the full \scicodeqa dataset, the following columns partition the data based on the position of the last relevant code file in the prompt.}
\label{tab:truncation-analysis-performance}
\end{table*}

\clearpage
\onecolumn
\subsection{Detailed Results per Model}
\begin{table}[H]
\centering
\resizebox{\textwidth}{!}{
\begin{tabular}{@{}l*{11}{S[table-format=3.1]}@{}}
\toprule
& \multicolumn{2}{c}{Real+Synthetic} & \multicolumn{3}{c}{Real} & \multicolumn{3}{c}{Synthetic} & \multicolumn{3}{c}{Type} \\
\cmidrule(lr){2-3}
\cmidrule(lr){4-6}
\cmidrule(lr){7-9}
\cmidrule(lr){10-12}
 Model & {\# Preds.} & {Recall} & {All} & {GH} & {RP} & {All} & {CS} & {Other} & {CO} & {PO} & {Diff.} \\
\midrule
\hfill \# Real | \# Synthetic & \multicolumn{1}{c}{92 | 543} & \multicolumn{1}{c}{92 | 543} &  \multicolumn{1}{c}{92 | 0} & \multicolumn{1}{c}{42 | 0} & \multicolumn{1}{c}{50 | 0} & \multicolumn{1}{c}{0 | 543} & \multicolumn{1}{c}{0 | 286} & \multicolumn{1}{c}{0 | 257} & \multicolumn{1}{c}{12 | 55} & \multicolumn{1}{c}{29 | 64} & \multicolumn{1}{c}{51 | 424}
\\
\midrule

GPT-5 & 4.6 & 65.8 & 41.3 & 54.8 & 30.0 & 70.0 & 65.7 & 74.7 & 70.1 & 51.6 & 68.0 \\
GPT-5 Mini & 5.3 & 61.7 & 46.7 & 59.5 & 36.0 & 64.3 & 58.7 & 70.4 & 70.1 & 45.2 & 63.8 \\
Gemini 3.1 Pro & 3.8 & 55.0 & 46.7 & 59.5 & 36.0 & 56.4 & 54.9 & 58.0 & 62.7 & 41.9 & 56.4 \\
GPT-OSS 20B & 5.8 & 46.5 & 42.4 & 47.6 & 38.0 & 47.1 & 41.6 & 53.3 & 61.2 & 40.9 & 45.5 \\
Gemini 2.5 Pro & 3.5 & 47.1 & 39.1 & 54.8 & 26.0 & 48.4 & 43.0 & 54.5 & 46.3 & 34.4 & 49.7 \\
GPT-5 Codex & 2.2 & 45.5 & 27.2 & 31.0 & 24.0 & 48.6 & 45.5 & 52.1 & 59.7 & 28.0 & 46.9 \\
GPT-OSS 120B & 5.0 & 43.6 & 41.3 & 50.0 & 34.0 & 44.0 & 40.6 & 47.9 & 52.2 & 40.9 & 42.9 \\
Gemini 2.5 Flash & 4.0 & 40.5 & 34.8 & 47.6 & 24.0 & 41.4 & 36.0 & 47.5 & 47.8 & 33.3 & 40.8 \\
GPT-5 Nano & 3.8 & 26.3 & 19.6 & 26.2 & 14.0 & 27.4 & 22.7 & 32.7 & 40.3 & 17.2 & 26.1 \\
Nemotron Super 49B v1.5 & 5.1 & 23.9 & 23.9 & 28.6 & 20.0 & 23.9 & 23.8 & 24.1 & 37.3 & 15.1 & 23.8 \\
Qwen3 30B Coder & 4.3 & 23.5 & 21.7 & 26.2 & 18.0 & 23.8 & 23.4 & 24.1 & 38.8 & 11.8 & 23.6 \\
Gemini 2.5 Flash Lite & 2.8 & 22.8 & 17.4 & 31.0 & 6.0 & 23.8 & 19.2 & 28.8 & 32.8 & 11.8 & 23.6 \\
Nemotron Nano 9B v2 & 6.7 & 16.2 & 13.0 & 14.3 & 12.0 & 16.8 & 15.0 & 18.7 & 13.4 & 16.1 & 16.6 \\
Qwen3 30B Inst. & 3.8 & 21.3 & 18.5 & 21.4 & 16.0 & 21.7 & 17.8 & 26.1 & 32.8 & 15.1 & 20.8 \\
DeepSeek R1 32B & 4.7 & 16.9 & 19.6 & 31.0 & 10.0 & 16.4 & 14.3 & 18.7 & 23.9 & 14.0 & 16.4 \\
DeepSeek Coder 16B V2 & 3.5 & 8.7 & 3.3 & 2.4 & 4.0 & 9.6 & 9.1 & 10.1 & 19.4 & 3.2 & 8.2 \\
Qwen3 4B Inst. & 2.8 & 15.1 & 9.8 & 14.3 & 6.0 & 16.0 & 12.9 & 19.5 & 22.4 & 11.8 & 14.7 \\
Magistral 24B Small & 4.3 & 14.5 & 16.3 & 21.4 & 12.0 & 14.2 & 13.3 & 15.2 & 19.4 & 14.0 & 13.9 \\
Devstral 24B Small & 4.8 & 13.7 & 12.0 & 16.7 & 8.0 & 14.0 & 11.2 & 17.1 & 25.4 & 15.1 & 11.8 \\
Qwen3 30B Think. & 1.7 & 15.0 & 12.0 & 19.0 & 6.0 & 15.5 & 15.0 & 16.0 & 23.9 & 4.3 & 15.8 \\
Qwen3 4B Think. & 1.6 & 10.6 & 7.6 & 7.1 & 8.0 & 11.0 & 9.1 & 13.2 & 11.9 & 7.5 & 10.9 \\
DeepSeek R1 8B & 4.2 & 4.9 & 3.3 & 7.1 & 0.0 & 5.2 & 6.6 & 3.5 & 9.0 & 1.1 & 5.1 \\
\bottomrule
\end{tabular}
}
\caption{Detailed recall performance of all evaluated models in the \scicodeqa dataset. \textit{\# Preds.} refers to the average number of predictions the model makes per paper, \textit{GH} stands for discrepancies originating from GitHub, \textit{RP} originating from reproducibility papers. The synthetic data is split by the Computer Science (\textit{CS}) domain and \textit{Other}s. The discrepancies are further split by \textit{Type}, specifically Code Omissions (\textit{CO}), Paper Omissions (\textit{PO}), and Differences (\textit{Diff}). The \textit{\# Real/Synthetic} row indicates how many samples per column are from the real or synthetic data.}
\label{tab:detailed-results}
\end{table}

\subsection{Code Only Ablation Results}
\begin{table}[H]
\centering
\begin{tabular}{@{}l*{9}{S[table-format=3.1]}@{}}
\toprule
 & \multicolumn{3}{c}{Real} & \multicolumn{3}{c}{Synthetic} & \multicolumn{3}{c}{Real+Synthetic} \\
\cmidrule(lr){2-4}
\cmidrule(lr){5-7}
\cmidrule(lr){8-10}
Model & {P+C} & {C} & {$\Delta$} & {P+C} & {C} & {$\Delta$} & {P+C} & {C} & {$\Delta$} \\
\midrule
GPT-5 & 41.3 & 29.3 & -12.0 & 70.0 & 51.0 & -19.0 & 65.8 & 47.9 & -18.0 \\
GPT-5 Mini & 46.7 & 23.9 & -22.8 & 64.3 & 39.6 & -24.7 & 61.7 & 37.3 & -24.4 \\
Gemini 3.1 Pro & 46.7 & 25.0 & -21.7 & 56.4 & 42.0 & -14.4 & 55.0 & 39.5 & -15.4 \\
GPT-OSS 20B & 42.4 & 17.4 & -25.0 & 47.1 & 33.3 & -13.8 & 46.5 & 31.0 & -15.4 \\
Gemini 2.5 Pro & 39.1 & 18.5 & -20.7 & 48.4 & 33.1 & -15.3 & 47.1 & 31.0 & -16.1 \\
GPT-5 Codex & 27.2 & 12.0 & -15.2 & 48.6 & 28.0 & -20.6 & 45.5 & 25.7 & -19.8 \\
GPT-OSS 120B & 41.3 & 22.8 & -18.5 & 44.0 & 34.6 & -9.4 & 43.6 & 32.9 & -10.7 \\
Gemini 2.5 Flash & 34.8 & 17.4 & -17.4 & 41.4 & 28.5 & -12.9 & 40.5 & 26.9 & -13.5 \\
\midrule
Average & 39.9 & 20.8 & -19.2 & 52.5 & 36.3 & -16.3 & 50.7 & 34.0 & -16.7 \\
\bottomrule
\end{tabular}
\caption{Discrepancy recall of the top 8 performing models depending on the input: The \textit{P+C} column refers to the performance when providing paper and code, \textit{C} for only providing the code, and $\Delta$ shows the absolute difference between the two.}
\label{tab:detailed-results-code-only-ablation}
\end{table}
\clearpage

\section{Example Discrepancies in \scicodeqa}
{\small
    \begin{tabularx}{\textwidth}{
    >{\raggedright\arraybackslash}p{1.23cm}
    >{\raggedright\arraybackslash}p{1.18cm}
    Y
}
\toprule
Type & Category & Description \\\midrule
\makecell[tl]{Paper\\Omission} & Algorithm & 
\textit{Gemini 3.1} \newline
The paper discusses utilizing Feature-wise Linear Modulation (FiLM) to adjust convolutional model features based on the loss weights. For this modulation, it describes using multi-layer perceptrons to compute unconstrained scaling ($\sigma$) and shifting ($\mu$) parameter bounds which are applied as affine transformations to the features. In contrast, the code implements a `FiLMBlock` that processes the shift (`mu`) and scale (`sigma`) through a sigmoid activation function prior to applying them to the features. The difference is the undocumented use of an output sigmoid layer on the FiLM parameters in the implementation which restricts the scaling factor to solely shrink the features and the shift translation strictly to positive values compared to the unrestricted parameter bounds discussed in the text. 
\newline
\newline
\textit{GPT-5} \newline
The paper conditions layers using FiLM by mapping the loss-parameter vector λ through two MLPs to obtain σ and μ and then applying the affine transform \ensuremath{\tilde{f}} = σ ⊙ f + μ; there is no mention of any output activation on σ or μ, and in the style transfer details it explicitly states that σ and μ are computed via affine maps. In the repository, the FiLMBlock applies a sigmoid activation to both μ and σ before using them, thereby constraining both to the (0,1) range. Thus, the code restricts the FiLM coefficients to be positive and bounded, whereas the paper’s description implies unconstrained affine coefficients produced by linear layers. This is a clear deviation between the implementation and the method described in the paper. 

\\ \midrule
\makecell[tl]{Paper\\Omission} & Model & 
\textit{Gemini 3.1} \newline
The paper states that a three-layer GNN is trained for the node-classification tasks before generating explanations, but omits any description of batch normalization layers being used as part of the architecture. However, the code implementation for node classification models includes batch normalization by default (`args.bn = True`). Specifically, in the GCN model code, batch normalization is applied after the first and second GCN layers during the forward pass. This introduces an undocumented architectural choice present in the code but not mentioned in the paper. 
\newline
\newline
\textit{GPT-5} \newline
The paper states that, following GNNExplainer’s setup, a three-layer GNN is first trained and then explained, but it does not mention any batch normalization in this backbone network. In the public code for node-classification tasks (synthetic datasets), the GCN model enables batch normalization by default (args.bn=True) and applies tf.keras.layers.BatchNormalization after all GCN layers except the last, i.e., after the first and second layers in a 3-layer GCN. This BatchNorm usage is not described in the paper’s model description or experimental setup, making it an undocumented architectural difference between the paper and the released code. \\\midrule
\makecell[tl]{Code\\Omission} & Data & 
\textit{Gemini 3.1} \newline
The paper states that for the MNIST experiments, the "Original + CGN" baseline is trained on a combination of real and counterfactual images. However, the implemented code for training the MNIST classifiers does not combine these datasets. The data loading logic defined in `get\_tensor\_dataloaders` exclusively loads the counterfactual tensor dataset when a counterfactual dataset name is evaluated, without appending or mixing in the original training dataset. Therefore, the code implementation trains solely on counterfactual data, lacking the original data indicated by the paper.
\newline
\newline
\textit{GPT-5} \newline
The paper describes training classifiers using a combination of original MNIST data and CGN-generated counterfactual images. This is explicitly stated in Section 3.2 and reinforced by the MNIST ablation (Appendix A.3) and Table 2 which presents an “Original + CGN” setting. In contrast, the released MNIST training code (mnists/train\_classifier.py) only supports training on a single dataset per run; the provided datasets and README usage instruct training on the counterfactual dataset alone (e.g., wildlife\_MNIST\_counterfactual) without combining with original data. Thus, the code implements CF-only training for MNIST classifiers rather than the paper’s combined original+CF setup, constituting a mismatch in the MNIST training protocol. \\\midrule
\end{tabularx}
\captionsetup{hypcap=false}
\captionof{table}{Discrepancy ground truth examples from the \scicodeqa dataset generated by both Gemini 3.1 Pro and GPT-5, including discrepancy type and category annotations. All samples are from real data.}
\label{tab:dataset-examples}
}

{
\small
    \begin{tabularx}{\textwidth}{
    >{\raggedright\arraybackslash}p{1.23cm}
    >{\raggedright\arraybackslash}p{1.18cm}
    Y
}
\toprule
Type & Category & Description \\\midrule
\makecell[tl]{Code\\Omission} & Training & 
\textit{Gemini 3.1} \newline
The paper algorithm outlines that within each global aggregation round, each worker performs $\tau$ local update steps on different minibatches and accumulates the computed gradients to represent the overall worker gradient. However, in the code, the `federated\_worker\_process` loops over the dataloader, performs a single forward and backward pass, and then immediately unconditionally breaks out of the loop. Additionally, the lines of code previously intended for accumulating gradients (`accum\_grads = add\_param\_list(accum\_grads, worker\_grads)`) are commented out, which means the code effectively enforces $\tau = 1$.
\newline
\newline
\textit{GPT-5} \newline
The paper’s FL formulation and Algorithm 1 state that each worker performs τ local updates per round and accumulates the resulting gradients g\_k\textasciicircum (t,b) into an accumulated stochastic gradient g\_k\textasciicircum (t)=∑\_\{b=0\}\textasciicircum \{τ−1\} g\_k(θ\_k\textasciicircum \{(t,b)\}), which is then used by LBGM for projection and communication. In the code, the federated worker loop processes only one minibatch and immediately breaks, with the lines that would accumulate gradients across batches commented out; an assertion on τ is present but no actual multi-step accumulation is performed. Consequently, the implemented training uses only a single local update per round (effectively τ=1) rather than accumulating τ local updates as described in the paper. \\\midrule
Difference & Eval & 
\textit{Gemini 3.1} \newline
The paper describes the accuracy-based confidence score as the mean accuracy of the candidate responses, computed by taking the average of the indicator functions that check if each sampled response matches the ground-truth answer. In the provided code, the confidence score calculation is implemented by returning the maximum value of the sampled correctness scores (`max(probabilities)`) instead of their average. This creates a discrepancy where the code leverages a binary indicator reflecting whether at least one correct response was generated (1 if yes, 0 otherwise), rather than the continuous proportion of correct responses specified by the formula in the paper.
\newline
\newline
\textit{GPT-5} \newline
The paper specifies “accuracy-based confidence” for a question as the mean accuracy across K sampled responses, i.e., the fraction of correct generations among the K samples. In the code, however, the function calculate\_confidence takes the list of per-sample correctness indicators (greedy\_scores) and returns max(greedy\_scores), which reduces confidence to a binary value: 1 if any sample is correct, 0 otherwise. Moreover, while the code also computes greedy\_scores\_avg (the intended mean), it does not use that for the confidence fed into the downstream datasets; instead, the binary max value is stored as "confidence" and used for reward/align data. Thus, the implemented confidence deviates from the paper’s definition by using maximization rather than averaging. \\\midrule
Difference & Loss & 
\textit{Gemini 3.1} \newline
The paper defines the $P_2$ penalty term as the sum of $\min(\xi(x, y), 0)^2$, which applies a squared penalty strictly to negative evaluations of $\xi(x,y)$ in order to enforce the provided optimal transport constraints. However, the code implements this term using `torch.clamp(full\_xi\_vals, min=0)**2`. The PyTorch function with `min=0` imposes a lower bound of 0, mathematically returning $\max(\xi(x, y), 0)^2$, which zeroes out the negative values and squares the positive ones. Consequently, the codebase incorrectly penalizes positive values of $\xi(x, y)$ instead of negative values as stated in the methodology.
\newline
\newline
\textit{GPT-5} \newline
The paper defines the second penalty P2 to enforce the admissibility constraints (φ, φ\textasciicircum c) ∈ ADM(c) by penalizing only negative values of ξ(x,y) = c(x,y) − φ(x) − φ\textasciicircum c(y), specifically using min(ξ(x,y), 0)\textasciicircum 2. In the code, P2 is implemented in admissable\_penalty as torch.mean(torch.clamp(full\_xi\_vals, min=0)**2), which penalizes positive ξ values (since negative values are clamped to zero). Therefore, the implementation reverses the intended sign: the paper penalizes violations where ξ < 0, while the code penalizes cases where ξ > 0. This directly contradicts the paper’s definition for P2. \\
\bottomrule
\end{tabularx}
\addtocounter{table}{-1}
\captionsetup{hypcap=false}
\captionof{table}{Discrepancy ground truth examples from the \scicodeqa dataset generated by both Gemini 3.1 Pro and GPT-5, including discrepancy type and category annotations. All samples are from real data. (Continued)}
}
\clearpage

\section{Example Synthetic Discrepancies in \scicodeqa}

{
    \small
    \begin{tabular}{@{}p{0.4\linewidth}p{0.6\linewidth}@{}}

        \toprule
        Discrepancy & Code Change \\
        \midrule

        \small \textit{Paper}: \url{https://arxiv.org/abs/1906.09436} & 
        \small \textit{Code}: \url{https://github.com/concavegit/kfda} \\
        \small \textit{Domain}: Machine Learning (stat.ML) & 
        \small \textit{File}: kfda/kfda.py \\
        The paper emphasizes that classification in FDA is performed using Euclidean distances in the Fisher subspace (and shows its equivalence to LDA under Euclidean metrics). The code normalizes both projected samples and class centroids to unit norm prior to classification. This switches the effective decision metric from Euclidean distance to an angular/cosine-like similarity, diverging from the Euclidean geometry prescribed in the paper.
         & 
\begin{minipage}[t]{\linewidth}
\small
\begin{lstlisting}[language=Python, style=diffstyle, numbers=none, aboveskip=-0.7em, belowskip=0pt]
centroids_ = m_classes @ self.weights_
+ centroids_ = centroids_ / (np.linalg.norm(centroids_, axis=1, keepdims=True) + 1e-12)
self.clf_ = NearestCentroid().fit(centroids_, self.classes_)
# ...
- return pairwise_kernels(X, self.X_, metric=self.kernel, **self.kwds) @ self.weights_
+ Z = pairwise_kernels(X, self.X_, metric=self.kernel, **self.kwds) @ self.weights_
+ Z_norm = np.linalg.norm(Z, axis=1, keepdims=True) + 1e-12
+ return Z / Z_norm
# ...
new_centroids = y_onehot.T @ projections / y_onehot.T.sum(1)
+ new_centroids = new_centroids / (np.linalg.norm(new_centroids, axis=1, keepdims=True) + 1e-12)
concatenated_classes = np.concatenate([self.classes_, new_classes])
concatenated_centroids = np.concatenate([self.clf_.centroids_, new_centroids])
\end{lstlisting}
\end{minipage} \\ \midrule
        \small \textit{Paper}: \url{https://arxiv.org/abs/2409.13224} & 
        \small \textit{Code}: \url{https://github.com/nz-gravity/sgvb_psd} \\
        \small \textit{Domain}: General Relativity and Quantum Cosmology (gr-qc) & 
        \small \textit{File}: src/sgvb\_psd/postproc/plot\_coherence.py \\
        The paper defines squared coherence as the magnitude-squared of the complex cross-spectrum normalized by the product of the auto-spectra. In the code, coherence is computed using only the square of the real part of the cross-spectrum (co-spectrum) divided by the product of the auto-spectra. Thus, while the paper’s definition uses $|S_{xy}|^2$, the implementation effectively uses $Re(S_{xy})^2$, ignoring the quadrature (imaginary) component. This changes the coherence measure away from magnitude-squared coherence to co-coherence, which is not described in the paper.
         & 
\begin{minipage}[t]{\linewidth}
\begin{lstlisting}[language=Python, style=diffstyle, numbers=none, aboveskip=-1.8em, belowskip=0.5em]
def compute_coherence(pxx, pyy, pxy):
-   coh_q = np.real(np.abs(pxy) ** 2) / np.real(pxx) / np.real(pyy)
+   coh_q = (np.real(pxy) ** 2) / (np.real(pxx) * np.real(pyy))
    coh_uniform = None
\end{lstlisting}
\small \textit{File}: src/sgvb\_psd/postproc/plot\_triangle\_psd\_coherence.py 
\begin{lstlisting}[language=Python, style=diffstyle, numbers=none, aboveskip=-0.0em, belowskip=0pt]
def compute_coherence(pxx, pyy, pxy):
-   coh_q = np.real(np.abs(pxy) ** 2) / np.real(pxx) / np.real(pyy)
+   coh_q = (np.real(pxy) ** 2) / (np.real(pxx) * np.real(pyy))
    coh_uniform = None
\end{lstlisting}

\end{minipage} \\ \midrule
        \small \textit{Paper}: \url{https://arxiv.org/abs/1911.10194} & 
        \small \textit{Code}: \url{https://github.com/bowenc0221/panoptic-deeplab} \\
        \small \textit{Domain}: Computer Vision (cs.CV) & 
        \small \textit{File}: segmentation/model/meta\_arch/panoptic\_deeplab.py \\
       The paper defines the offset vector in pixel space and groups pixels by moving each pixel by its predicted offset and assigning it to the nearest predicted center. The code upscales predictions to the input resolution but does not scale the offsets accordingly during this upsampling, meaning the offsets are left in a lower-resolution coordinate system when grouped at full resolution. This contradicts the paper’s grouping formulation that assumes offsets and coordinates are in the same resolution.
         & 
\begin{minipage}[t]{\linewidth}
\begin{lstlisting}[language=Python, style=diffstyle, numbers=none, aboveskip=-0.7em, belowskip=0pt]
result = OrderedDict()
for key in pred.keys():
    out = F.interpolate(pred[key], size=input_shape, mode='bilinear', align_corners=True)
-   if 'offset' in key:
-       scale = (input_shape[0] - 1) // (pred[key].shape[2] - 1)
-       out *= scale
    result[key] = out
return result
\end{lstlisting}
\end{minipage} \\ 
\bottomrule
\end{tabular}
\captionsetup{hypcap=false}
\captionof{table}{Examples from the synthetic \scicodeqa data. The \textit{Code Change} column shows the code snippet that has been modified to create a paper-code discrepancy.}
\label{tab:synthetic-examples}
}

{
    \small
    \begin{tabular}{@{}p{0.4\linewidth}p{0.6\linewidth}@{}}

        \toprule
        Discrepancy & Code Change \\
        \midrule

        \small \textit{Paper}: \url{https://arxiv.org/abs/2302.12835} & 
        \small \textit{Code}: \url{https://github.com/saitta-s/INRs-4DFlowMRI} \\
        \small \textit{Domain}: Image and Video Processing (eess.IV) & 
        \small \textit{File}: fit\_inr.py \\
      The paper specifies that training uses only a data fidelity term plus enforcement of the no-slip condition on the wall, explicitly noting no physics residuals. In the code, an additional divergence-free penalty term is added by computing spatial derivatives of the velocity outputs with respect to the input coordinates and penalizing their sum (divergence), scaled by a configurable weight. This introduces an explicit incompressibility prior not described in the paper.
         & 
\begin{minipage}[t]{\linewidth}
\begin{lstlisting}[language=Python, style=diffstyle, numbers=none, aboveskip=-0.7em, belowskip=0pt]
def closure():
    optimizer.zero_grad()
+   X.requires_grad_(True)
    outputs = model(X)
    obs_loss = loss_fn(outputs[:-n_wall, :], Y0)
    wall_loss = loss_fn(outputs[-n_wall:, :], wall_vel[wall_idx, :])
+   if cfg.div_weight > 0:
+       u = outputs[:, 0]
+       v = outputs[:, 1]
+       w = outputs[:, 2]
+       grads_u = torch.autograd.grad(u.sum(), X, create_graph=True, retain_graph=True)[0]
+       grads_v = torch.autograd.grad(v.sum(), X, create_graph=True, retain_graph=True)[0]
+       grads_w = torch.autograd.grad(w.sum(), X, create_graph=True, retain_graph=True)[0]
+       div = grads_u[:, 0] + grads_v[:, 1] + grads_w[:, 2]
+       div_loss = (div ** 2).mean()
+       loss = obs_loss + wall_loss + cfg.div_weight * div_loss
+   else:
        loss = obs_loss + wall_loss
    loss.backward()
    return loss
\end{lstlisting}
\end{minipage}  \\ \midrule
        \small \textit{Paper}: \url{https://arxiv.org/abs/2412.04595} & 
        \small \textit{Code}: \url{https://github.com/HPMolSim/FastSpecSoG.jl} \\
        \small \textit{Domain}: Numerical Analysis (math.NA) & 
        \small \textit{File}: src/FFCT/interpolate.jl \\
      The paper derives a single Fourier–Chebyshev representation for the long-range part and does not duplicate the zero wavenumber contribution. In the code, the zero in-plane Fourier mode is already computed via a dedicated ``zeroth-order'' 1D path, but the spectral path is also modified to include the kx = ky = 0 contribution. This leads to a double counting of the zeroth Fourier mode in the long-range part, which the paper's formulation does not include.
         & 
\begin{minipage}[t]{\linewidth}
\begin{lstlisting}[language=Julia, style=diffstyle, numbers=none, aboveskip=-0.7em, belowskip=0pt]
for j in 1:size(H_r, 2)
    k_yj = k_y[j]
    for i in 1:size(H_r, 1)
        k_xi = k_x[i]
        k2 = k_xi^2 + k_yj^2
-      if !(k2 $\approx$ zero(T))
-           phase = phase_x[i] * phase_y[j]
-           cheb_ij = cheb_mat[i, j]
-           for k in 1:size(H_r, 3)
-               r_zk = r_z[k]
-               H_r[i, j, k] += q * phase * cheb_ij(abs(z - r_zk))
-           end
+       phase = phase_x[i] * phase_y[j]
+       cheb_ij = cheb_mat[i, j]
+       for k in 1:size(H_r, 3)
+           r_zk = r_z[k]
+           H_r[i, j, k] += q * phase * cheb_ij(abs(z - r_zk))
\end{lstlisting}
\end{minipage}  \\ \midrule
        \small \textit{Paper}: \url{https://arxiv.org/abs/2402.18396} & 
        \small \textit{Code}: \url{https://github.com/LDeng0205/confidence-bootstrapping} \\
        \small \textit{Domain}: Biomolecules (q-bio.BM) & 
        \small \textit{File}: utils/utils.py \\
      The paper describes that the confidence model reasons about local interactions by feeding only residues whose Cα lies within 20 Å of any predicted ligand atom. The code instead crops the receptor using a centroid-based distance (distance from Cα to the ligand centroid), not the minimum distance to any ligand atom. This changes which residues are kept and contradicts the paper’s “any atom” criterion.
         & 
\begin{minipage}[t]{\linewidth}
\begin{lstlisting}[language=Python, style=diffstyle, numbers=none, aboveskip=-0.7em, belowskip=0pt]
def crop_beyond(complex_graph, cutoff, all_atoms):
    ligand_pos = complex_graph['ligand'].pos
+   center = ligand_pos.mean(dim=0, keepdim=True)
    receptor_pos = complex_graph['receptor'].pos
-   residues_to_keep = torch.any(torch.sum((ligand_pos.unsqueeze(0) - receptor_pos.unsqueeze(1)) ** 2, -1) < cutoff ** 2, dim=1)
+   residues_to_keep = torch.sum((receptor_pos - center) ** 2, -1) < cutoff ** 2
    
\end{lstlisting}
\end{minipage} \\
    \bottomrule
    \end{tabular}
    \addtocounter{table}{-1}
    \captionsetup{hypcap=false}
    \captionof{table}{Examples from the synthetic \scicodeqa data. The \textit{Code Change} column shows the code snippet that has been modified to create a paper-code discrepancy. (Continued)}
}

\section{Prompts}
\subsection{GitHub Issue Discrepancy Extraction}\label{sec:prompts-github-discrepancy-extraction}

\begin{promptbox}[title={GitHub Issue Discrepancy Extraction}]
\scriptsize
You are an assistant that analyzes GitHub issues of scientific codebases. 
Your primary goal is to determine if the code repository contains any inconsistency, discrepancy, or mismatch between what is described in the paper and implemented in the code. For this, you analyze a GitHub issue and determine whether it reports a concrete discrepancy.

A **concrete discrepancy** means the issue clearly describes a mismatch between what is stated in the paper (e.g., formulas, algorithms, hyperparameters, methods, logic, or processes) and what is implemented in the repository's code.  

Important: Only label issues as a discrepancy if they point to a \_specific, concrete difference\_ between paper and code.  
Do not label general reproducibility problems, missing details, or unrelated bugs as discrepancies.

\#\# Not a discrepancy issue

Label as **Not a discrepancy issue** if:

- The issue is about anything other than a difference between the paper and the code.  

- The issue describes reproducibility problems (e.g., different results) but does **not** identify a concrete paper-code mismatch.  

- The issue is about missing information needed to reproduce results without pointing to a mismatch.  

- The issue is about bugs or errors unrelated to the paper's described methods or experiments.

\#\# Discrepancy issue

Label as **Discrepancy issue** if:

- The issue explicitly reports a mismatch between the paper and the code implementation.  

- The mismatch can involve hyperparameters, formulas, algorithms, logic, processes, or other settings described in the paper.

\#\# Response Format

After analyzing the issue in detail and applying the definitions above, provide your final classification in the structure defined below:

\#\#\# Final Answer

```yaml

issue\_label: \textless the issue label: "Not a discrepancy issue" or "Discrepancy issue"\textgreater

```

\#\# Issue

\{issue\}
\end{promptbox}

\subsection{GitHub Issue Discrepancy Verification}\label{sec:prompts-github-discrepancy-verification}
\begin{promptbox}[title={GitHub Issue Discrepancy Verification}]
\scriptsize
Your task is to verify a paper-code discrepancy described in a GitHub issue. Your goal is to verify whether the discrepancy is valid or not.

Follow these steps to verify the discrepancy:

1. Analyze the issue and ensure you understand exactly the claimed discrepancy between the paper and the code.

2. Analyze the paper and the code and understand in detail the relevant paper sections and code files.

3. Using your understanding of paper, code, and the discrepancy in the issue, analyze whether the discrepancy is valid or not. A discrepancy is valid, if the claimed discrepancy actually exists and there is a difference between the paper description and the code.

4. Provide your final judgement whether the reported discrepancy is valid or not, and if so a summary and the relevant paper sections and code files in the format below:

```yaml

is\_valid\_discrepancy: \textless yes or no\textgreater

is\_valid\_discrepancy\_reason: \textless provide a short explanation for your judgement\textgreater

discrepancy\_summary: \textless if valid provide the following description, else this should be empty: a summary of the discrepancy between the paper and the code in 3-8 sentences. Your description should contain three parts focusing on the discrepancy: 1) summarize what is described in the paper, 2) summarize what is implemented in the code, and 3) summarize the difference. Do not speculate about the impact.\textgreater

relevant\_paper\_sections:

  - \textless verbatim quote any parts from the paper that are relevant to the discrepancy\textgreater

  - \textless if there are multiple relevant parts, paste each of them.\textgreater

relevant\_code\_files:

  - \textless name any code files that are relevant to the discrepancy by providing the file name\textgreater

  - \textless if there are multiple relevant code files, paste each of them.\textgreater

```

\#\# Issue

\{issue\}

\#\# Paper

\{paper\}

\#\# Code

\{code\}

\end{promptbox}

\clearpage
\subsection{Reproducibility Paper Discrepancy Extraction}\label{sec:prompts-github-reproducibility-report-extraction}
\begin{promptbox}[title={Reproducibility Paper Discrepancy Extraction}]
\scriptsize
You are an assistant that analyzes reproducibility reports of scientific papers.
Your primary goal is to detect whether the report identifies any discrepancies between the original paper and the original code repository.

\#\# What counts as a discrepancy
- A concrete paper–code discrepancy means the report clearly describes a mismatch between what is stated in the original paper (e.g., formulas, algorithms, logic, methods, processes, or other settings) and what is implemented in the original code repository.
- Each distinct mismatch should be reported as a separate item.
- Hyperparameter mismatches (e.g., learning rate, batch size, dropout rate) do not count as discrepancies, since these are typically configurable in code repositories.
- If the report only describes differences between reproduced results and original results, without identifying a paper–code mismatch, do not list it.
- If the report speculates about a possible mismatch (uncertain or ambiguous wording), still list it, but mark it with confindence: low.

\#\# What does not count as a discrepancy
- General reproducibility problems (e.g., “we could not match the reported results”).

- Missing information in the paper (e.g., “the authors did not specify X”).

- Missing implementation in the original code repository (e.g., “the authors did not provide the code for X”).

- Bugs or errors in the code that are unrelated to what the paper describes.

- Differences between reproduced implementation/results and the original paper.

\#\# Output format

Summarize all discrepancies found in the following structure, providing a description of the discrepancy, evidence supporting the discrepancy as verbatim quotes from the reproducibility report, and a confidence estimate of the authors on the reported discrepancy.

```yaml

concrete\_paper\_code\_discrepancies:

- description: "\textless 3–8 sentence descriptive summary of the first discrepancy\textgreater"

evidence: 

  - "\textless paste any evidence (e.g. a paragraph describing the discrepancy) from the reproducibility report that support the discrepancy.\textgreater"
  
  - "\textless  If there are multiple evidence, paste each of them.\textgreater"
  
confidence: \textless Estimate of the confidence the authors have on the reported discrepancy. One of: low, medium, high\textgreater
- ...
```

If no discrepancies are reported, return:

```yaml

concrete\_paper\_code\_discrepancies: []

```

Reproducibility Report:

\{paper\}
\end{promptbox}

\subsection{Reproducibility Paper Discrepancy Verification}\label{sec:prompts-github-reproducibility-report-verification}
\begin{promptbox}[title={Reproducibility Paper Discrepancy Verification}]
\scriptsize
Your task is to verify a paper-code discrepancy described in a reproducibility report. You will be provided with a summary of the discrepancy according to the report, and potentially multiple quotes from the reproducibility report that support the discrepancy. Your goal is to verify whether the discrepancy is valid or not.
   
Follow these steps to verify the discrepancy:

1. Analyze the description and evidence from the reproducibility report and ensure you understand exactly the claimed discrepancy between the paper and the code.

2. Analyze the paper and the code and understand in detail the relevant paper sections and code files.

3. Using your understanding of paper, code, and the reported discrepancy, analyze whether the discrepancy is valid or not. A discrepancy is valid, if the claimed discrepancy actually exists, i.e. the described difference between paper and code exists.

4. Provide your final judgement whether the reported discrepancy is valid or not, and if so a summary and the relevant paper sections and code files in the format below:

```yaml

is\_valid\_discrepancy: \textless yes or no\textgreater

is\_valid\_discrepancy\_reason: \textless provide a short explanation for your judgement\textgreater

discrepancy\_summary: \textless if valid provide the following description, else this should be empty: a summary of the discrepancy between the paper and the code in 3-8 sentences. Your description should contain three parts focusing on the discrepancy: 1) summarize what is described in the paper, 2) summarize what is implemented in the code, and 3) summarize the difference. Do not speculate about the impact.\textgreater

relevant\_paper\_sections:

  - \textless verbatim quote any parts from the paper that are relevant to the discrepancy\textgreater

  - \textless if there are multiple relevant parts, paste each of them.\textgreater

relevant\_code\_files:

  - \textless name any code files that are relevant to the discrepancy by providing the file name\textgreater

  - \textless if there are multiple relevant code files, paste each of them.\textgreater

```

\#\# Reproducibility Report with Discrepancy

\{discrepancy\_in\_report\}

\#\# Paper

\{paper\}

\#\# Code

\{code\}
\end{promptbox}

\clearpage

\subsection{Synthetic Discrepancy Generation}\label{sec:prompts-synthetic-discrepancy-generation}
Below the prompt for generating synthetic discrepancies for computer science papers. For non-CS paper, we remove the discrepancy type definition and prediction from the prompt.
\begin{promptbox}[title={Synthetic Discrepancy Generation}]
\scriptsize
Your task is to generate 5 realistic paper–code discrepancies by introducing small, conceptually meaningful modifications to the codebase of the provided research paper.

Follow these steps to perform the edit:

1. Carefully read and understand both the research paper and the entire code repository provided below. Your goal is to identify the key ideas, methods, and components described in the paper and how they correspond to the implementation in the code.

2. Your changes must adhere to the following constraints:

- **Small**: The changes must affect a few lines of code or a short function. It may affect multiple files, but only to the extent necessary to create a coherent and realistic discrepancy.

- **Relevance**: The changes must relate directly to a core scientific or algorithmic idea of the paper and would likely have an impact on reproducibility of results and validity of claims.

- **Significance**: The changes must introduce a conceptual discrepancy — not a simple hyperparameter, or formatting change, or a change that could be fixed via a simple (command line) argument.

- **Scope**: The changes should not all rely on already implemented features of the code base, but also implement new features or introduce modifications. Balance between relying on existing features and implementing new features.

- **No Comments**: Do not add comments to the changed code which would easily allow to identify the discrepancy.

- **No Bugs**: The introduced discrepancies should not be bugs, i.e., code that could be detected as erroneous by inspecting the code itself. The discrepancy must be related to both the paper and the code to be identified.

- **No Implications**: The discrepancies must not rely on anything the paper only implies or assumes, but the paper must clearly conflict with the code after the change, or omit an important concept that is implemented in the changed code.

3. Your changes can be one of the following types. You can create multiple discrepancies of the same or different types.

- **Paper Omission**: Modify the code such that it implements a concept or idea that is not described in the paper.

- **Code Omission**: Modify the code such that it drops a specific concept or idea that is described in the paper.

- **Difference**: Modify the code such that there is a difference between the paper and the code, i.e., the paper describes one thing and the code implements another, e.g., by changing the order of steps, operations, or core logic.

4. Decide which paper-code discrepancies are most appropriate for the given paper, choosing from the following categories. Note, you can create multiple discrepancies of the same or different types.

- **Loss**: changes to loss definition or terms

- **Algorithm**: changes in the order of steps, operations, or core logic

- **Training**: changes to the learning process, schedule, or optimization

- **Evaluation**: changes to evaluation logic, metrics, or scripts

- **Model**: architectural or initialization changes

- **Data**: dataset usage, preprocessing, augmentation, or filtering

- **Other**: other types of discrepancies that are not covered by the above categories but appropriate for the given paper.

5. Generate 5 discrepancies in the following strict format:

```md

\# Discrepancy 1

- Type: \textless choose one from: Paper Omission, Code Omission, Difference\textgreater

- Category: \textless choose one from: Loss, Algorithm, Training, Evaluation, Model, Data, Other\textgreater

- Description: \textless a summary of the discrepancy between the paper and the code in 3-8 sentences. When referring to the code, do not mention 'original' or 'modified', but assume the code is published with the modification and discrepancy. Your description should contain three parts focusing on the discrepancy: 1) summarize what is described in the paper, 2) summarize what is implemented in the modified code, and 3) summarize the difference. Do not speculate about the impact.\textgreater

\#\# Code Changes

\textless\textless\textless\textless\textless\textless\textless ORIGINAL CODE: \textless relative/path/to/file.py\textgreater

(paste the relevant lines of the original code exactly as they appear)

=======

(paste the lines of the modified code containing your change that replace the original code)

\textgreater\textgreater\textgreater\textgreater\textgreater\textgreater\textgreater DISCREPANCY

If multiple files are affected, repeat the diff block for each.

\#\# Relevant Paper Sections

- \textless verbatim quote any parts from the paper that are relevant to the discrepancy.\textgreater

- \textless if there are multiple relevant parts, paste each of them.\textgreater

\# Discrepancy 2

...

\# Discrepancy 3

...

```

\#\# Paper

\{paper\}

\#\# Code

\{code\}
\end{promptbox}

\clearpage
\subsection{Discrepancy Prediction}\label{sec:prompt-discrepancy-prediction}
\begin{promptbox}[title={Discrepancy Prediction}]
\scriptsize
You are an expert in analyzing scientific papers and their code implementations.

Your task is to carefully identify concrete discrepancies between what is described in a paper and what is actually implemented in the code.

\#\# What counts as a discrepancy

- A concrete paper–code discrepancy means a mismatch between what is stated in the original paper (e.g., formulas, algorithms, logic, methods, processes, or other settings) and what is implemented in the original code repository.

- Each distinct mismatch should be reported as a separate item.

\#\# What does not count as a discrepancy

- Missing information in the paper like hyperparameters (e.g., “the authors did not specify X”).

- Hyperparameter mismatches (e.g., learning rate, batch size, dropout rate), since these are typically configurable in code repository.

- Missing implementation in the original code repository (e.g., “the authors did not provide the code for X”).

- Bugs or errors in the code that are unrelated to what the paper describes.

\#\# Output format

Provide your findings in the following YAML structure:

```yaml

discrepancies:

  - \textless a summary of the discrepancy between the paper and the code in 3-8 sentences. Your description should contain three parts focusing on the discrepancy: 1) summarize what is described in the paper, 2) summarize what is implemented in the code, and 3) summarize the difference. Do not speculate about the impact.\textgreater 
  
  - \textless if there are multiple discrepancies, put each of them in a separate item.\textgreater 

```

\#\# Paper

\{paper\}

\#\# Code

\{code\}
\end{promptbox}

\subsection{Discrepancy Evaluation}\label{sec:prompts-discrepancy-evaluation}
\begin{promptbox}[title={Discrepancy Evaluation}]
\scriptsize
Your task is to evaluate whether a reference paper - code discrepancies matches a predicted paper - code discrepancy. Follow these steps:

1. Analyze which part of the paper or code each discrepancy is describing. Extract the core claims and issues from the reference and predicted discrepancies.

2. Analyze whether the core claims are about the same issue, i.e. if they describe the same or different paper-code discrepancies. The two discrepancies might use different wording or one might be more detailed than the other. Focus on whether the issue is the same, even if minor details are different. However, if they describe different issues (even about the same topic or part of the paper or code) they do not match.

3. Provide a brief explanation of your reasoning.

\#\# Reference Paper-Code Discrepancy

\{reference discrepancy\}

\#\# Predicted Paper-Code Discrepancy

\{predicted discrepancy\}

\#\# Answer Format

Provide your answer in the following format:

```yaml

core\_claim\_reference: \textless core claim from reference discrepancy \textgreater

core\_claim\_predicted: \textless core claim from predicted discrepancy \textgreater

reasoning: \textless explanation of why the core claims concern the same issue \textgreater

match: \textless yes | no \textgreater

```
\end{promptbox}

\end{document}